\documentclass[journal]{IEEEtran}
\usepackage{multirow}
\usepackage{array}
\usepackage{makecell}   
\usepackage{graphicx}
\usepackage{amsfonts}   
\usepackage{amsmath}    
\usepackage{cite}      
\usepackage{booktabs}   
\usepackage[table,xcdraw]{xcolor}   
\usepackage{bm}         
\usepackage{color}      
\usepackage{algorithm}
\usepackage{algorithmic}
\usepackage{setspace}
\usepackage{multirow}
\usepackage{tikz}
\usepackage{pgfplots}
\usepackage{amssymb}
\usepackage{adjustbox}
\usepackage{amsmath}
\usepackage{float}

\usepackage{xcolor}

\ifCLASSINFOpdf
\else
\fi

\begin{document}

% paper title
% Titles are generally capitalized except for words such as a, an, and, as,
% at, but, by, for, in, nor, of, on, or, the, to and up, which are usually
% not capitalized unless they are the first or last word of the title.
% Linebreaks \\ can be used within to get better formatting as desired.
% Do not put math or special symbols in the title.
\title{OV-NeRF: Open-vocabulary Neural Radiance Fields with Vision and Language Foundation Models \\ for 3D Semantic Understanding }

% author names and IEEE memberships
% note positions of commas and nonbreaking spaces ( ~ ) LaTeX will not break
% a structure at a ~ so this keeps an author's name from being broken across
% two lines.
% use \thanks{} to gain access to the first footnote area
% a separate \thanks must be used for each paragraph as LaTeX2e's \thanks
% was not built to handle multiple paragraphs
%

\author{Guibiao Liao,
        Kaichen Zhou, 
        Zhenyu Bao, 
        Kanglin Liu, 
        and Qing Li
        % ,~\IEEEmembership{Member,~IEEE,}
        % ,~\IEEEmembership{Senior Member,~IEEE,}
        % ,~\IEEEmembership{Fellow,~IEEE} 
% \thanks{This work was supported by (\textit{Corresponding author: Qing Li.})}
\thanks{Guibiao Liao and Zhenyu Bao are with the School of Electronic and Computer Engineering, Peking University, Shenzhen, China, and also with Pengcheng Laboratory, Shenzhen, China (Email: gbliao269@gmail.com, zybao@pku.edu.cn).}
\thanks{Kaichen Zhou is with the Department of Computer Science, University of Oxford, Oxfordshire, England (Email: rui.zhou@cs.ox.ac.uk).}
\thanks{Kanglin Liu and Qing Li are with Pengcheng Laboratory, Shenzhen, China (Email: max.liu.426@gmail.com, lqing900205@gmail.com).}% <-this % stops a space 
\thanks{\textit{Corresponding author: Qing Li; Kanglin Liu.}}% <-this % stops a space 
% \thanks{Manuscript received April 19, 2005; revised August 26, 2015.}
}

% The paper headers
% \markboth{Journal of \LaTeX\ Class Files,~Vol.~14, No.~8, month~year}%
\markboth{IEEE Transactions on Circuits and Systems for Video Technology}
{Shell \MakeLowercase{\textit{et al.}}: Bare Demo of IEEEtran.cls for IEEE Journals}

\maketitle

\begin{abstract}
The development of Neural Radiance Fields (NeRFs) has provided a potent representation for encapsulating the geometric and appearance characteristics of 3D scenes. 
Enhancing the capabilities of NeRFs in open-vocabulary 3D semantic perception tasks has been a recent focus. 
However, current methods that extract semantics directly from Contrastive Language-Image Pretraining (CLIP) for semantic field learning encounter difficulties due to noisy and view-inconsistent semantics provided by CLIP. 
To tackle these limitations, we propose OV-NeRF, which exploits the potential of pre-trained vision and language foundation models to enhance semantic field learning through proposed single-view and cross-view strategies. 
First, from the \textit{single-view} perspective, we introduce Region Semantic Ranking (RSR) regularization by leveraging 2D mask proposals derived from Segment Anything (SAM) to rectify the noisy semantics of each training view, facilitating accurate semantic field learning. 
Second, from the \textit{cross-view} perspective, we propose a Cross-view Self-enhancement (CSE) strategy to address the challenge raised by view-inconsistent semantics. 
Rather than invariably utilizing the 2D inconsistent semantics from CLIP, CSE leverages the 3D consistent semantics generated from the well-trained semantic field itself for semantic field training, aiming to reduce ambiguity and enhance overall semantic consistency across different views. 
Extensive experiments validate our OV-NeRF outperforms current state-of-the-art methods, achieving a significant improvement of 20.31\% and 18.42\% in mIoU metric on Replica and ScanNet, respectively. 
Furthermore, our approach exhibits consistent superior results across various CLIP configurations, further verifying its robustness. 
Codes are available at: https://github.com/pcl3dv/OV-NeRF. 
\end{abstract}

\begin{IEEEkeywords}
Neural radiance field, open-vocabulary, vision and language foundation models, cross-view self-enhancement.  
\end{IEEEkeywords}

\IEEEpeerreviewmaketitle

\IEEEpubid{\begin{minipage}{\textwidth}\ \centering
    Copyright \copyright 20xx IEEE. Personal use of this material is permitted. \\However, permission to use this material for any other purposes must be obtained from the IEEE by sending an email to pubs-permissions@ieee.org. 
\end{minipage}}

\IEEEpubidadjcol

\section{Introduction}
% Neural Radiance Fields (NeRFs) 
% 3D Open-vocabulary semantic understanding 
\IEEEPARstart{N}{eural} Radiance Fields (NeRFs) have emerged as a promising representation method for capturing complex real-world 3D scenes. They have shown their proficiency in rendering high-quality novel views \cite{nerf, mipnerf, zipnerf, tensorf, instantngp, nerfreview, zhou2023dynpoint, li2023representing, guo2024depth}. However, achieving a comprehensive 3D semantic understanding \cite{robotnavigation, jaritz2019multi, feng2020deep, conceptfusion, rong2021active, yin2023dcnet, shi2023temporal} remains a challenging problem. 
To realize this goal, an intuitive manner is leveraging manually annotated multi-view semantic labels to supervise NeRF to learn a semantic field for semantic rendering \cite{semanticNeRF}. 
Nevertheless, the resource-intensive nature of manual annotation hinders its practical application in real-world NeRF-based 3D semantic understanding. 
Recently, owing to the impressive performance in open-vocabulary image understanding, Contrastive Language-Image Pre-Training (CLIP) \cite{OpenAICLIP, OpenCLIP}, serving as a vision and language foundation model (VLM), has shown intriguing potential for open-vocabulary 2D semantic understanding \cite{LSeg, FCCLIP, regionclip, zhou2022extract, CLIPSelf, Segclip, xu2023side, liang2023open, zhang2023simple}, eliminating the need for annotated semantic labels. 
In light of this, leveraging unlabeled images, corresponding language descriptions, and CLIP models to pursue NeRF-based open-vocabulary 3D semantic understanding, emerges as a promising research area.

Taking into account the insights mentioned above, the studies conducted by \cite{LERF, 3DOVS} utilize a pre-trained CLIP model and open vocabularies to attain 3D semantic comprehension with NeRF. 
Specifically, 
LERF \cite{LERF} constructs a semantic field jointly with NeRF and distills pixel-level semantic features of each training view from the CLIP image encoder into the semantic field. In this way, LERF generates 3D open-vocabulary relevancy maps with text queries. 
Subsequently, Liu et. al. \cite{3DOVS} similarly align pixel-level CLIP features with a designed semantic field. Additionally, they introduce a relevancy map alignment loss to enhance semantic supervision from CLIP.

\begin{figure*}
\centering
\includegraphics[width=.85\linewidth]{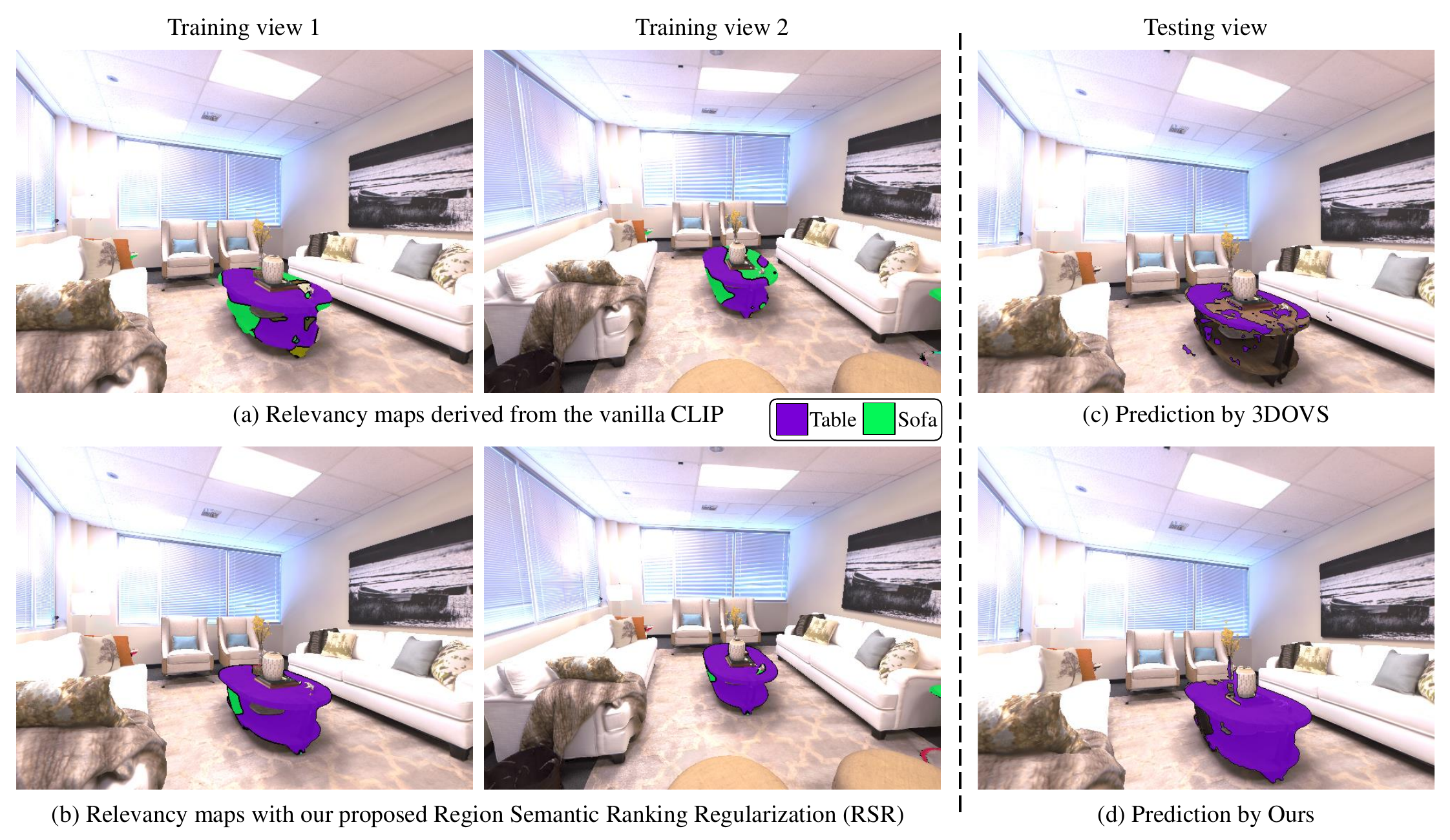}
\caption{Visualization of relevancy maps with the text query ``Table". 
As shown in (a), the two relevancy maps of different views derived from the original CLIP exhibit much \textit{coarseness} and \textit{view inconsistency}. However, 3DOVS \cite{3DOVS} directly leverages these relevancy maps for semantic field learning, leading to inferior rendering results, as exhibited in (c).
In contrast, the proposed Region Semantic Ranking regularization significantly improves the quality of relevancy maps as illustrated in (b), yielding the precise result in (d). 
}
\label{fig:vis_fig1}
\end{figure*}

\IEEEpubidadjcol

However, some limitations may curtail their effectiveness in achieving precise 3D semantic understanding.
\textbf{1) Coarse Relevancy Maps.} 
As depicted in the training view 1 of the first row in Fig. \ref{fig:vis_fig1}, it is evident that the relevancy map derived from the CLIP image encoder exhibits coarse and noisy semantics (purple and green are mixed in Table), particularly along the object boundary. 
This limitation arises from the fact that vanilla CLIP is primarily designed for image classification, thus exhibiting shortcomings in tasks requiring dense prediction, especially the object-level integrity \cite{CLIPSelf}. 
Yet, the aforementioned methods straightforwardly utilize these coarse maps to train semantic fields, struggling to obtain accurate 3D segmentation results. 
\textbf{2) View-inconsistent Relevancy Maps.} 
As illustrated in different views of the first row in Fig. \ref{fig:vis_fig1}, inconsistent semantics occur across multi-view relevancy maps. 
This phenomenon arises from the inherent challenge faced by the 2D CLIP model in maintaining consistent object identities across diverse perspectives. 
Nevertheless, current methods rely on these view-inconsistent relevance maps to train the semantic field, thereby posing a challenge to achieving multi-view consistent rendering results. 
Hence, the principal challenge in employing the CLIP model for NeRF-based open-vocabulary semantic understanding lies in how to effectively optimize the semantic field with imperfect relevancy maps.

To address these limitations, we present \textbf{OV-NeRF}, an innovative approach for accurate NeRF-based open-vocabulary 3D semantic understanding, considering both \textit{single-view} and \textit{cross-view} perspectives. 
From the \textit{single-view} perspective, it is noteworthy that the "Table" region predominantly showcases correct semantic classification, albeit lacking fine boundaries. 
To overcome this limitation, we draw inspiration from the powerful zero-shot region proposal extraction capability of vision foundation model Segment Anything (SAM) \cite{SAM}, and introduce Region Semantic Ranking (RSR) regularization to improve the accuracy of single view relevancy map. 
RSR incorporates regional hints naturally from SAM, providing region-level semantic regularization to enhance the precision of boundary results, as depicted in (b) of Fig. \ref{fig:vis_fig1}. 
Then, from the \textit{cross-view} perspective, we propose a Cross-view Self-enhancement (CSE) strategy to tackle the view inconsistency of relevancy maps.  
Informed by the inherent 3D consistency of NeRFs, the CSE leverages the 3D consistent relevancy maps generated from the trained semantic field. These maps, further refined with the RSR strategy (refer to (e) of Fig. \ref{fig:vis_rsr_cse}), are employed for semantic field training instead of the view-inconsistent relevancy maps obtained from CLIP. 
Additionally, CSE produces unseen semantic novel views, incorporating the RSR strategy, to provide additional informative cues in the same region across different views (refer to Fig. \ref{fig:vis_novel_view}), further enhancing the view consistency. 
In this way, multi-view semantic ambiguities from the CLIP model can be mitigated, thereby improving semantic consistency across different views.

To verify the 3D semantic understanding capability of our approach, we conduct experiments on two indoor scene datasets: Replica \cite{replica} and ScanNet \cite{scannet}. 
Extensive experiments demonstrate that the proposed RSR and CSE strategies collectively enable our OV-NeRF to significantly outperform existing state-of-the-art methods, achieving a remarkable improvement of 20.31\% and 18.42\% in terms of the mIoU metric on Replica and ScanNet, respectively. Additionally, both qualitative and quantitative ablation results affirm the effectiveness of each proposed strategy.  
The main contributions of this work are summarized as follows. 

\begin{itemize}
    \item We propose OV-NeRF, an innovative approach harnessing language and image knowledge from pre-trained vision and language foundation models, i.e., CLIP and SAM, to achieve precise NeRF-based open-vocabulary 3D semantic understanding.
    \item We introduce Region Semantic Ranking (RSR) regularization to address the noisy semantics issue and produce accurate single-view semantics for semantic field training. 
    \item Additionally, we propose Cross-view Self-enhancement (CSE) to reduce ambiguity and ensure semantic consistency enhancement across different views. 
    \item Extensive experiments demonstrate that our approach outperforms state-of-the-art methods in open-vocabulary 3D scene segmentation. 
    Moreover, our method consistently exhibits superior performance across various CLIP configurations, validating its generalizability. 
\end{itemize} 

The structure of the remainder of this paper is outlined as follows.
Section II provides a comprehensive review and discussion of related works. 
Subsequently, in Section III, we elucidate our approach in detail. 
In Section IV, we illustrate the conducted experiments and provide extensive analysis.
Finally, Section V concludes this work.

\section{Related Work}
\subsection{NeRFs for Semantic Modeling}
To represent 3D scenes, Neural Radiance Field (NeRF) \cite{nerf, nerfreview, Nerfies, pixelnerf, mipnerf360, Nerfren, wang2022nerfcap, Nerfplayer, wei2023depth, Structnerf} employs a continuous multi-layer perceptron (MLP) that learns the color and density of continuous coordinates. Through this coordinate-based MLP, the radiance and volume density can be queried by input position and view direction vectors, enabling the synthesis of photo-realistic novel views. 
To enhance the rendering quality of novel view synthesis, Mip-nerf \cite{mipnerf} introduced cone tracing as an alternative to traditional point-based ray tracing to mitigate aliasing issues. Additionally, ZipNeRF \cite{zipnerf} presented an anti-aliased grid-based method aimed at improving the overall performance of the radiance field. 
To boost the training and rendering efficiency of NeRFs, Instant-NGP \cite{instantngp} minimized the expense associated with neural primitives through an adaptable input encoding. This facilitated the use of a more compact network without compromising quality. TensoRF \cite{tensorf} represented radiance fields as 4D tensors, decomposing them into efficient low-rank tensor components using CANDECOMP/PARAFAC (CP) \cite{cp_decomposition} and vector-matrix (VM) decompositions. 
However, these approaches face limitations in addressing 3D semantic understanding, i.e., semantic modeling of 3D scenes.

To model semantics based on NeRF for 3D scenes, semantic-NeRF \cite{semanticNeRF} 
established a semantic field, and leveraged annotated semantic labels to supervise NeRF, facilitating the synthesis of semantic masks from novel views. 
Nevertheless, this modeling approach, which introduces a large number of annotated labels, is non-trivial and expensive. 
Subsequently, 
Tschernezki et. al. \cite{n3f} introduced a teacher-student framework, which distilled image features obtained from the frozen DINO \cite{DINO} model into the trainable semantic field for semantic feature learning.  
Fan et. al. \cite{nerfsos} also utilized the DINO model to extract 2D features, and they distilled the feature correspondence into the semantic field in a conservative learning manner. 
Unlike previous approaches that used visual features from DINO, several researchers have recently explored how to utilize textual descriptions in combination with CLIP models \cite{OpenAICLIP, OpenCLIP} to achieve NeRF-based open-vocabulary 3D semantic understanding. 
For instance, DFF \cite{DFF} delved into embedding pixel-aligned feature vectors from LSeg \cite{LSeg} for semantic feature field optimization. 
LERF \cite{LERF} extended the CLIP's feature alignment idea to a scale-conditioned feature field, which was supervised by using multi-scale and pixel-level CLIP features derived from the CLIP image encoder. 
The recent effort 3DOVS \cite{3DOVS} similarly optimized a semantic feature field by using pixel-level CLIP features. In addition, 3DOVS introduced a Relevancy-Distribution Alignment loss to promote semantic learning from CLIP models. 
However, as illustrated in Fig. \ref{fig:vis_fig1}, the above methods straightforwardly utilize these noisy and view-inconsistent semantics from CLIP to train the semantic field, yielding imprecise NeRF-based open-vocabulary 3D semantic understanding. 

% On the contrary, our approach overcomes these challenges by leveraging vision and language foundation models from both \textit{single-view} and \textit{cross-view} perspectives to train the semantic field, improving the accuracy and view consistency of semantics for NeRF-based 3D semantic understanding. 

In contrast to previous methods, we propose RSR and CSE to tackle the issues of coarse semantics and view inconsistency, respectively. RSR imposes region-level semantic regularization, developed based on SAM, to rectify coarse boundaries. CSE leverages refined 3D consistent semantics derived from a trained NeRF to provide effective view-consistent supervision, enhancing semantic consistency across diverse views.

\begin{figure*}
\centering
\includegraphics[width=\linewidth]{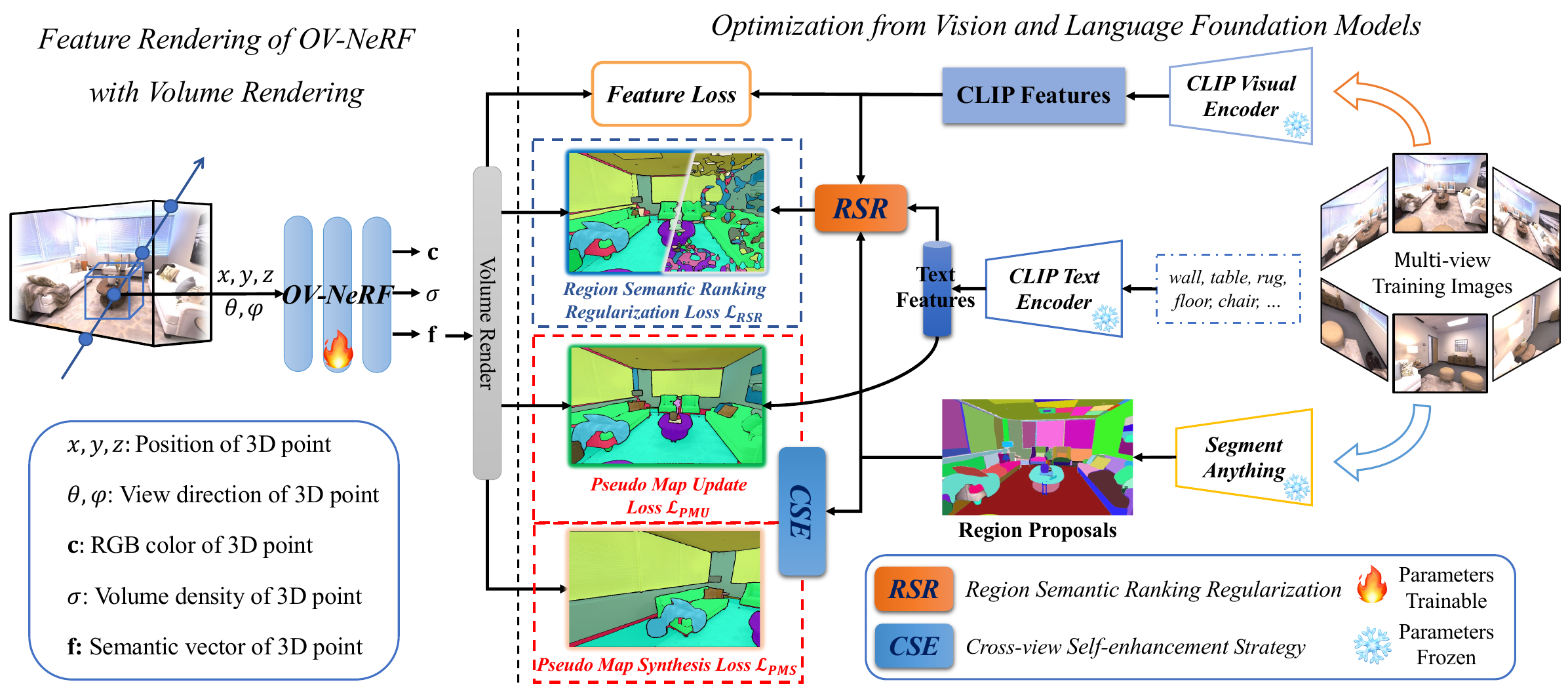}
\caption{
\textbf{Overview of OV-NeRF Optimization: } 
\textit{Left}: OV-NeRF feature rendering process. OV-NeRF represents a semantic field of 3D volumes using a trainable MLP network (Section \ref{Overview_framework}). 
\textit{Right}: Optimization of the trainable OV-NeRF by utilizing vision and language foundation models. 
Initially, multi-view training images undergo processing in the CLIP visual encoder to extract image features. Concurrently, text embeddings are obtained through the CLIP text encoder. 
% Then, the Segment Anything model is employed to generate region proposals over the corresponding images. 
To optimize OV-NeRF, we propose two key strategies: \textbf{Region Semantic Ranking (RSR)} regularization and \textbf{Cross-view Self-enhancement (CSE)} strategy, along with the incorporation of the CLIP feature loss. 
Specifically, the Segment Anything model is employed to produce region proposals over the corresponding images. Then, utilizing pre-computed CLIP features and SAM's region proposals, our RSR generates the precise relevancy map (blue border) to supervise OV-NeRF (Section \ref{RSR_design}), instead of using the original noisy relevancy map (gray border) derived from the CLIP model. 
Furthermore, after training OV-NeRF for several epochs, we leverage the rendered pseudo outputs obtained from OV-NeRF, encompassing both training views (green border) and unseen novel views (orange border), for cross-view self-enhancement supervision (Section \ref{CSE_design}). 
}
\label{fig:overview}
\end{figure*}

\subsection{Vision and Language Foundation Models}
Pre-trained Vision and Language Models (VLMs) \cite{flamingo, OpenAICLIP, OpenCLIP} have emerged as a potent paradigm in computer science. These models undergo pre-training on extensive datasets, endowing them with the capacity to excel in a variety of downstream tasks, either in a zero-shot manner or with fine-tuning. 
For instance, recent works such as Contrastive Language-Image Pre-Training (CLIP) \cite{OpenAICLIP, OpenCLIP} and Segment Anything Model (SAM) \cite{SAM} have garnered considerable attention due to their remarkable achievements in zero-shot image classification and zero-shot class-agnostic segmentation, respectively.
CLIP combines a visual image encoder and a text encoder, and leverages an image-text contrastive learning strategy to learn associations between images and corresponding text descriptions.
This paradigm utilizes large-scale web-crawled image-text pair data to pre-train the image encoder and text encoder, facilitating the image encoder to achieve open-world image classification \cite{regionclip, OpenAICLIP, OpenCLIP, LSeg, FCCLIP} based solely on text data. 
SAM, serves as a strong vision foundation model, consisting of an image encoder, a prompt encoder, and a mask decoder. 
Segment Anything (SAM) undergoes training on an extensive dataset comprising 11 million images and 1 billion masks. This substantial corpus of annotated data empowers SAM to yield exemplary {class-agnostic region proposals} that exhibit robust performance across diverse distributions \cite{samtrack, sammedical, samhq}. 

While CLIP and SAM demonstrate impressive capabilities in 2D open-world image understanding, their direct applicability to 3D tasks presents challenges. These challenges arise from the scarcity of extensive 3D annotations and large-scale 3D-text data. 
In this paper, we harness the capability of CLIP and SAM for NeRF-based 3D open-vocabulary segmentation.

\section{Methodology}
First, the preliminaries and overview of the proposed OV-NeRF are briefly presented in Section \ref{Overview_framework}. 
Following this, a detailed introduction of the proposed Region Semantic Ranking (RSR) regularization is provided in Section \ref{RSR_design}, and the Cross-view Self-enhancement (CSE) strategy is elaborated upon in Section \ref{CSE_design}. 
Finally, we present the training loss functions in Section \ref{Loss_design}.

\subsection{Preliminaries and Overview} \label{Overview_framework}
Our approach is designed to establish the semantic perception of 3D scenes represented by the Neural Radiance Field (NeRF) \cite{nerf}, which takes a set of $N$ posed views $I = \{ I_i \in \mathbb{R}^{H \times W \times 3} | i \in \{ 1, ..., N \} \}$ as input. 
% Our algorithm is designed to establish the semantic perception based on NeRF for an indoor scenario by leveraging a set of $N$ input views $I = \{ I_i \in \mathbb{R}^{H \times W \times 3} | i \in \{ 1, ..., N \}\}$ with known camera poses. 
 Concretely, given the input posed image $I_i$, a set of text descriptions, our objective is to generate novel semantic views of specified viewpoints. 
To achieve this goal, we harness the image-text knowledge from the pre-trained vision and language foundation model CLIP \cite{OpenAICLIP, OpenCLIP}, and construct a semantic field in which each location is associated with a corresponding CLIP feature representing its semantic information.

% \textbf{Scene Representation and Rendering.}
Neural Radiance Field (NeRF) \cite{nerf} represents a 3D scene through an implicit, neural volumetric representation using multi-layer perceptrons (MLPs) $\Gamma_{\theta}: (\mathbf{x}, \mathbf{d}) \to (\mathbf{c}, \sigma)$, where $\theta$ indicates the learnable network parameters. This is a continuous function that outputs an RGB color $\mathbf{c} \in \mathbb{R}^3$ and a volume density $\sigma \in \mathbb{R}$ given a 3D position $\mathbf{x} \in \mathbb{R}^3$ and a view direction $\mathbf{d}$. 
In world coordinates, the input posed image can be conceptualized as sets of viewing rays. 
Following the principles of volume rendering, a ray $\mathbf{r}$ emitted from the camera center $\mathbf{o}$ is parameterized in 3D space as $\mathbf{r}(t) = \mathbf{o} + t\mathbf{d}$, and the color $\hat{\mathbf{C}}(\mathbf{r})$ alone this ray is formulated as:
\begin{equation}
\hat{\mathbf{C}}(\mathbf{r}) = \sum_{k} T_k \alpha_k \mathbf{c}_k \in \mathbb{R}^3, 
\end{equation}
where $\mathbf{c}_k$ is the color corresponding to the $k^{th}$ sampled 3D point along the ray. 
$T_k$ is the transmittance probability, which is defined as: $T_k = \prod_{j=0}^{k-1} (1 - \alpha_j)$.
Here, $\alpha_k$ is the opacity of the sampled 3D point $k$, and it is formulated as: $\alpha_k = 1 - \mathrm{exp}(- \delta_{k} \sigma_{k})$, where $\delta_{k} = t_{k+1} - t_{k}$ denotes the distance between two neighboring sampled points. 

To render the semantic feature $\hat{\mathbf{S}}(\mathbf{r})$ of each ray $\mathbf{r}$, we follow \cite{3DOVS} to introduce an additional MLP branch to achieve this: 
\begin{equation}
\hat{\mathbf{F}}(\mathbf{r}) = \sum_{k} T_k \alpha_k \mathbf{f}_k \in \mathbb{R}^D, 
\end{equation}
where $\mathbf{f}_k$ is the semantic vector of the $k^{th}$ sampled points of the ray. 
$D$ is the feature dimension of the semantic vector, which is 512 in \cite{LERF, 3DOVS}.
To obtain the rendered segmentation logits $\hat{\mathbf{S}}$ of each ray $\mathbf{r}$, we first leverage a set of text descriptions of $M$ classes and the CLIP text encoder to generate textual features $t_c \in \mathbb{R}^{M \times D}$. Subsequently, the rendered segmentation logits $\hat{\mathbf{S}}$ can be obtained by calculating the cosine similarities between the textual features $t_c$ and rendered semantic feature $\hat{\mathbf{F}}(\mathbf{r})$. This process can be formulated as: 
\begin{equation}
\hat{\mathbf{S}}(\mathbf{r}) = \mathrm{cos} \langle t_c, \hat{\mathbf{F}}(\mathbf{r}) \rangle. \label{eq_relevancy_map_clip}
\end{equation}

% loss: photometric RGB + feature + relevancy map + DINO
To optimize the semantic field, a random sampling of rays from all training images is adopted in the form of the batch. Four supervision losses are used to train the semantic field, including photometric RGB loss, relevancy feature loss, relevancy map loss, and DINO feature loss. The DINO feature $\hat{\mathbf{D}}(\mathbf{r})$ of each ray $\mathbf{r}$ is computed for regularization as in \cite{LERF, 3DOVS}. 
The photometric RGB loss is formulated as the L2 distance between rendered and ground truth RGB color: $\mathcal{L}_{color} = \sum_{\mathbf{r}} ||\hat{\mathbf{C}}(\mathbf{r}) - {\mathbf{C}}(\mathbf{r})||^2_2$. 
The relevancy feature loss is the cosine distance between the rendered semantic features and CLIP features: $\mathcal{L}_{feat} = \sum_{\mathbf{r}} - \mathrm{cos} \langle \hat{\mathbf{F}}(\mathbf{r}), \mathcal{S}(\mathbf{r}) {\mathbf{F}}(\mathbf{r}) \rangle $. 
The relevancy map loss adopts the cross-entropy loss $\mathcal{L}_{ce}$ between rendered segmentation logits and relevancy maps derived from CLIP features: $\mathcal{L}_{map} = \sum_{\mathbf{r}}  \mathcal{L}_{ce}( \hat{\mathbf{S}}(\mathbf{r}), {\mathbf{S}}(\mathbf{r}) ) $. 
The DINO feature loss is the FDA loss $\mathcal{L}_{fda}$ \cite{3DOVS} between rendered DINO features and DINO features: $\mathcal{L}_{dino} = \sum_{\mathbf{r}} \mathcal{L}_{fda} (\hat{\mathbf{D}}(\mathbf{r}), {\mathbf{D}}(\mathbf{r})) $. 
Here, ${\mathbf{F}}(\mathbf{r})$ is the pre-computed multi-scale CLIP relevancy feature and $\mathcal{S}(\mathbf{r})$ is the selection vector for suitable scale selection as defined in \cite{3DOVS}. 
${\mathbf{S}}(\mathbf{r})$ is the pre-computed relevancy map by using the ${\mathbf{F}}(\mathbf{r})$, Equation \eqref{eq_relevancy_map_clip} and the argmax function.  ${\mathbf{D}}(\mathbf{r})$ is the pre-computed DINO feature provided by DINO \cite{DINO}. 
Consequently, the supervision loss for the scene optimization can be formulated as: 
\begin{equation}
    \mathcal{L} = \mathcal{L}_{color} + \mathcal{L}_{feat} + \mathcal{L}_{map} + \mathcal{L}_{dino}. \label{eq_3dovs_supervision}
\end{equation}

However, though the semantic field can be constructed through the aforementioned process, the semantic relevancy maps from CLIP models are often noisy and view inconsistent, as depicted in Fig. \ref{fig:vis_fig1}. 
Directly employing the coarse and view-inconsistent relevancy maps might not provide precise 3D consistent segment results. Therefore, in this work, we aim to construct more precise and view-consistent semantic relevancy maps for optimization. 

To fulfill this objective, we propose OV-NeRF, as depicted in Fig. \ref{fig:overview}, which mainly consists of two key components: Region Semantic Ranking (RSR) regularization and Cross-view Self-enhancement (CSE). 

First, multi-view images undergo feature extraction in the CLIP visual encoder, and text embeddings are derived from the CLIP text encoder. Meanwhile. the Segment Anything model is utilized to generate region proposals for the corresponding images. 
Second, RSR utilizes pre-computed CLIP features and SAM's region proposals to improve the quality of object integrity and boundaries, boosting the single-view accuracy of relevancy maps for optimization (Section \ref{RSR_design}). 
Third, CSE leverages the rendered pseudo outputs obtained from the learned neural radiance field to enhance semantic consistency across multiple views (Section \ref{CSE_design}). 
Details of each strategy will be expounded upon in the subsequent sections.

\subsection{Region Semantic Ranking Regularization (RSR)} \label{RSR_design}
\begin{figure*}
\centering
\includegraphics[width=\linewidth]{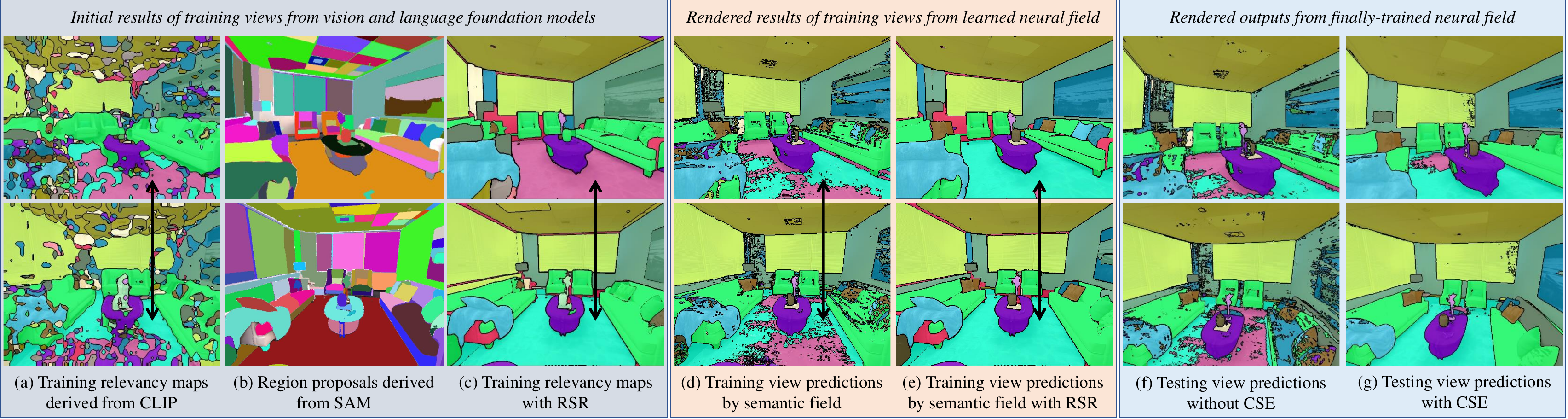}
\caption{
Visualization of relevancy maps. The first and second rows represent different views. 
(a) $\sim$ (e) are obtained from the training views, while (f) $\sim$ (g) are acquired from the testing views. For more details refer to Section \ref{CSE_design}.
}
\label{fig:vis_rsr_cse}
\end{figure*}

Although current methods \cite{LERF, 3DOVS} can achieve the association between 3D points and CLIP features via semantic field learning, the derived relevancy maps from CLIP models \cite{OpenAICLIP, OpenCLIP} encounter challenges such as coarse boundaries, leading to imprecise results of novel views, as shown in Fig. \ref{fig:vis_fig1}. 
To address this limitation, we aim to improve semantic field learning by training with more accurate single-view semantic relevancy maps. 
As illustrated in (a) of Fig. \ref{fig:vis_rsr_cse}, we note that each object region is dominated by correct semantic classification, such as the \textit{blinds} and \textit{table}. This implies that the region-level semantic regularization exhibits reliable classification information. 
Motivated by this, we introduce \textbf{Region Semantic Ranking} regularization, namely \textbf{RSR}, capitalizing on well-grouped region proposals provided by the powerful vision foundation model Segment Anything (SAM) \cite{SAM}.

\begin{algorithm}[!t]
    \setstretch{1.}
    \caption{Employing Region Semantic Ranking Regularization of an image}
    \label{alg_RSR}
      \begin{algorithmic}
        \STATE \textbf{Input:} Training RGB image $I_i$, Text description $T$, 
        CLIP image encoder $E_i$, CLIP text encoder $E_t$, SAM model $E_{sam}$
        \STATE  \textbf{Output:} Training relevancy map $\tilde{\mathbf{S}}_i$ 
        \STATE \textbf{Initialize:} $\tilde{\mathbf{S}}_i$ = zeros(1, $H$, $W$)
        
        ${\mathbf{F}}_i$ = $E_{i}$($I_i$) 
        
        $t_c$ = $E_{t}$($T$)         

        ${\mathbf{S}}_i = \mathrm{argmax} (\mathrm{cos} \langle t_c, {\mathbf{F}}_i \rangle )$
        
        $\mathcal{R}_i$ = $E_{sam}$($I_i$) 
        
        \FOR {$k=1$ to length($\mathcal{R}_i$)} \do 
    
            \begin{onehalfspacing}

              \STATE ${\mathbf{S}}_i^k$ = $\mathcal{R}_i^k$ * ${\mathbf{S}}_i$ 
              % \textcolor{blue}{ /* this is a comment */ }

              \STATE $\tilde{\mathbf{S}}_i^k$ = $\mathcal{R}_i^k$ * $\tilde{\mathbf{S}}_i$
              
              \STATE ${\mathbf{S}}_{i}^{k_{l}}$ = sort(${\mathbf{S}}_i^k$, descending)'[1] 
              
              \STATE $\tilde{\mathbf{S}}_i^k$ = ${\mathbf{S}}_{i}^{k_{l}}$
              
            \end{onehalfspacing}
    
        \ENDFOR

        \RETURN $\tilde{\mathbf{S}}_i$ 
        
      \end{algorithmic}
\end{algorithm}

% As shown in (b) of Fig. \ref{fig:vis_rsr_cse}, we pre-process the training images with pre-trained SAM to generate region proposals without semantic meaning. 
Specifically, as shown in (b) of Fig. \ref{fig:vis_rsr_cse}, we generate region proposals without semantic information for an entire training image $I_i$ by employing SAM's automatic mask generator. This process is achieved by querying SAM with uniformly sampling points on the images and three masks are obtained for each query point. Subsequently, based on the default setting, these masks of low confidence scores are filtered, and nearly identical masks are deduplicated using non-maximal suppression, yielding a collection of region mask proposals $\mathcal{R}_i = \{ \mathcal{R}_i^{j} \}_{j=1}^{n}$. Here, $n$ denotes the number of masks. 
Meanwhile, the relevancy map ${\mathbf{S}_i}$ is obtained by using a set of text descriptions and the CLIP image feature: ${\mathbf{S}}_i = \mathrm{argmax} (\mathrm{cos} \langle t_c, {\mathbf{F}}_i \rangle ) $, where $t_c$ denotes the text embedding produced from the CLIP text encoder and ${\mathbf{F}}_i$ is the image feature generated from CLIP image encoder. 
For the $k$-$th$ region proposal $\mathcal{R}_i^{k}$ of the relevancy map ${\mathbf{S}}_i$, we rank the semantic classes in descending order based on the pixel quantity of all classes and we can obtain the top one class ${\mathbf{S}}_{i}^{k_{l}}$. Here, $l$ indicates the $l$-$th$ class. 
Subsequently, the semantic of the region $\mathcal{R}_i^{k}$ is regularized as ${\mathbf{S}}_{i}^{k_{l}}$ by filtering out other classes. 
All region proposals of the training view $I_i$ are regularized in the same way. 
The whole process of Region Semantic Ranking Regularization is defined as $\mathbf{\Psi}$ and can be formulated as: 
\begin{equation}
\tilde{\mathbf{S}}_i = \mathbf{\Psi} ( {\mathbf{S}}_i ). 
\label{eq_RSR}
\end{equation}
The pseudo-code of RSR is presented in Alg. \ref{alg_RSR}. 
With region semantic ranking regularization, we can produce more precise training relevancy maps as exhibited in (c) of Fig. \ref{fig:vis_rsr_cse}. 
Therefore, we leverage these training relevancy maps with RSR instead of the original relevancy maps from CLIP to optimize the semantic field, which is defined as: 
\begin{equation}
\mathcal{L}_{RSR} = \sum_{\mathbf{r}} \mathcal{L}_{ce} (\hat{\mathbf{S}}(\mathbf{r}), \tilde{\mathbf{S}}(\mathbf{r})). 
\end{equation}
Note that, the RSR process $\mathbf{\Psi}$ is performed offline without introducing extra training time. 
During training, we only employ our region semantic ranking regularization loss $\mathcal{L}_{RSR}$ to replace the original relevancy maps loss to promote semantic field learning for precise rendering results.

\subsection{Cross-view Self-enhancement (CSE)} \label{CSE_design}
While the region semantic ranking regularization improves the single-view relevancy map accuracy, cross-view semantic consistency is still not guaranteed. 
As illustrated in (a) and (c) of Fig. \ref{fig:vis_rsr_cse}, the areas indicated by the arrows exhibit different semantic classes, specifically the \textit{floor} and the \textit{rug}. This limitation arises because the 2D CLIP model inherently struggles to maintain consistent object identities across diverse perspectives.
Consequently, conducting this deficient view-consistent learning leads to ambiguous results. 
Therefore, how to address this issue and inform our approach to the semantic consistency constraints of a scene is also a crucial task. 
Inspired by the inherent 3D consistency of NeRFs, we aim to leverage the view-consistent rendered outputs derived from the well-learned semantic field to enhance cross-view semantic consistency. 
To accomplish this, we propose a \textbf{Cross-view Self-enhancement (CSE)} strategy that consists of two parts: iterative pseudo map update and novel pseudo map synthesis.

\textbf{Iterative pseudo map update.} 
Specifically, after training the semantic field for a few iterations, we take the poses of training views $I = \{ I_i \in \mathbb{R}^{H \times W \times 3} | i \in \{ 1, ..., N \}\}$ as input for our learned semantic field and render the corresponding semantic relevancy maps as shown in (d) of Fig. \ref{fig:vis_rsr_cse}. 
Then, we further regularize these relevancy maps with our RSR to generate refined pseudo relevancy maps $\{ \bar{\mathbf{S}}_i \}_{i=1}^N$. 
The above update process of the pseudo relevancy maps is defined as $\mathbf{\Phi}$:
% and can be formulated as:  
\begin{equation}
\bar{\mathbf{S}}_i = \mathbf{\Phi} ( \hat{\mathbf{S}}_i ),  
\label{eq_RSR}
\end{equation}
where $\hat{\mathbf{S}}_i$ is the prediction obtained from the learned semantic field. 
As illustrated in (e) of Fig. \ref{fig:vis_rsr_cse}, we can observe that the areas indicated by the arrows show semantic consistency across these pseudo relevancy maps. 
Thus, we replace the relevancy maps provided by CLIP with our updated pseudo relevancy maps for semantic field learning. In this way, the semantic field is optimized by sampling random rays from all updated pseudo relevancy maps and minimizing the rendered and updated pseudo pixel. This process is calculated as: 
\begin{equation}
\mathcal{L}_{PMU} = \sum_{\mathbf{r}} \mathcal{L}_{ce} (\hat{\mathbf{S}}(\mathbf{r}), \bar{\mathbf{S}}(\mathbf{r}) ).  
\end{equation}
Moreover, as training progresses, these pseudo maps are updated at a regular interval $\mathcal{N}$. 
In this way, with more view-consistent supervision, the learned semantic field can be enhanced to generate more view-consistent pseudo maps. This iterative process, in turn, promotes view-consistent semantic learning within the semantic field, resulting in cross-view consistency enhancement as demonstrated in (g) of Fig. \ref{fig:vis_rsr_cse}. 

\begin{figure}
\centering
\includegraphics[width=\linewidth]{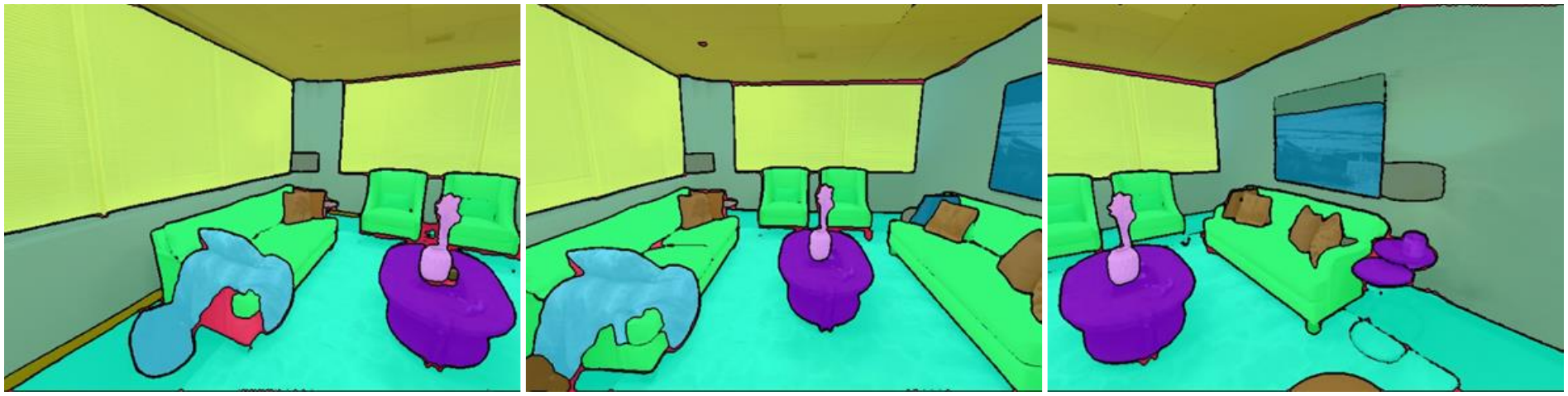}
\caption{
Visual examples of novel semantic pseudo map synthesis across different views. These novel pseudo maps can show the main structure of the scene, providing additional information for semantic field learning. 
}
\label{fig:vis_novel_view}
\end{figure}

\textbf{Novel pseudo map synthesis.}
In addition to semantic pseudo map update for training views, we consider utilizing unseen views to further enhance cross-view semantic consistency. 
Concretely, apart from conducting the above pseudo map update of training views, we simultaneously generate novel pseudo maps of novel viewpoints. 
Besides, these novel maps are also regularized with our RSR to obtain the fine edge and integral object, producing a set of novel pseudo relevancy maps $\{ {\mathbf{S}}_i^{+} \}_{i=1}^{N^{+}}$. Here, ${N^{+}}$ is the number of generated novel views, which is set as 60 in this work. 
As shown in Fig. \ref{fig:vis_novel_view}, the generated novel pseudo maps can capture the main structure of the scene, introducing informative semantic cues to enhance view consistency. 
Therefore, we introduce these additional pseudo maps to supervise the novel rendered relevancy maps $\{ \hat{\mathbf{S}}_i^{+} \}_{i=1}^{N^{+}}$ derived from the semantic field in a per-ray manner, which is formulated as: 
% This process can be formulated as: 
\begin{equation}
\mathcal{L}_{PMS} = \sum_{\mathbf{r}^{+}} \mathcal{L}_{ce} (\hat{\mathbf{S}}^{+}(\mathbf{r}^{+}), {\mathbf{S}}^{+}(\mathbf{r}^{+}) ), 
\end{equation}
where $\mathbf{r}^{+}$ indicates the sampled ray from the novel pseudo maps. 
With the help of these extended maps, the coherence of the 3D scene can be effectively enhanced, improving the cross-view semantic consistency. Therefore, our cross-view self-enhancement loss $\mathcal{L}_{CSE}$ can be calculated as:
\begin{equation}
\mathcal{L}_{CSE} = \mathcal{L}_{PMU} + \mathcal{L}_{PMS}. 
\end{equation} 
In summary, with the proposed Cross-view Self-enhancement (CSE), the ambiguity can be reduced, leading to more accurate multi-view rendering results, as verified in the experiments.

\subsection{Overall Training Loss} \label{Loss_design}
Our approach is conducted in an end-to-end training manner, consisting of two stages. 
{Stage I: at the beginning, we leverage the region semantic ranking regularization loss $\mathcal{L}_{RSR}$ to train the semantic field with a few iterations $\mathcal{T}$:}
% In the first stage, we leverage the region semantic ranking regularization loss $\mathcal{L}_{RSR}$ to train our model. This process is defined as: 
\begin{equation}
    \mathcal{L}_{Total} = \mathcal{L}_{color} + \mathcal{L}_{feat} + \mathcal{L}_{dino} + \mathcal{L}_{RSR}. 
\end{equation}
{Stage II: after training the semantic field a few iterations $\mathcal{T}$, we perform the cross-view self-enhancement loss $\mathcal{L}_{CSE}$ to replace the $\mathcal{L}_{RSR}$ for cross-view semantic enhancement: }
% After training a few iterations $\mathcal{T}$, we perform the cross-view self-enhancement loss $\mathcal{L}_{CSE}$ to replace the $\mathcal{L}_{RSR}$ for cross-view semantic enhancement, which can be formulated as: 
\begin{equation}
    \mathcal{L}_{Total} = \mathcal{L}_{color} + \mathcal{L}_{feat} + \mathcal{L}_{dino} + \mathcal{L}_{CSE}. 
\end{equation}

\begin{table*}[t]
\renewcommand{\arraystretch}{1.1}
\caption{Performance comparison with other methods on segmentation results of novel views from the Replica dataset when initialized with vanilla CLIP models (refer to Section \ref{section:comparison_SOTA}). Best results are highlighted as \textbf{bold}. }
\label{table:SOTA_Replica}
\centering
\small
\begin{adjustbox} {width=0.63\linewidth}
\begin{tabular}{ c c ||  p{0.9cm}<{\centering} p{0.9cm}<{\centering} p{0.9cm}<{\centering} | p{0.9cm}<{\centering}  p{0.9cm}<{\centering} p{0.9cm}<{\centering} }
\Xhline{3\arrayrulewidth}
\multirow{3}{*}{Dataset} & \multirow{3}{*}{Metric} & \multicolumn{3}{c|}{CLIP-based} & \multicolumn{3}{c}{OpenCLIP-based} \\ 
~ & ~  & {LERF} & {3DOVS} & {Ours}  & {LERF} & {3DOVS} & {Ours}\\ 
~ & ~  & \cite{LERF} & \cite{3DOVS} &   & \cite{LERF} & \cite{3DOVS} &  \\
\hline\hline
\multirow{2}{*}{Room0}   
& mIoU $\uparrow$  & 7.615  & 7.733  & \textbf{28.029} & 13.366 & 13.240  & \textbf{39.152} \\
& mAcc $\uparrow$  & 35.346 & 36.667 & \textbf{71.768} & 41.047 & 41.240  & \textbf{84.951} \\
\hline
\multirow{2}{*}{Room1} 
& mIoU $\uparrow$  & 14.043 & 16.016 & \textbf{29.802} & 12.002 & 12.300 & \textbf{32.968} \\
& mAcc $\uparrow$  & 30.218 & 35.957 & \textbf{68.755} & 28.741 & 29.286 & \textbf{74.510} \\
\hline
\multirow{2}{*}{Room2}   
& mIoU $\uparrow$  & 1.853  & 2.109  & \textbf{23.343} & 9.277  & 9.426  & \textbf{29.403} \\
& mAcc $\uparrow$  & 12.136 & 13.313 & \textbf{77.469} & 32.791 & 33.083 & \textbf{81.239} \\
\hline
\multirow{2}{*}{Office0} 
& mIoU $\uparrow$  & 3.257  & 3.793  & \textbf{15.805} & 3.623  & 3.781  & \textbf{10.416} \\
& mAcc $\uparrow$  & 4.724  & 6.271  & \textbf{54.617} & 13.874 & 14.131 & \textbf{23.993} \\
\hline
\multirow{2}{*}{Office1}   
& mIoU $\uparrow$  & 0.728  & 0.811  & \textbf{17.153} & 1.484  & 1.397  & \textbf{20.493} \\
& mAcc $\uparrow$  & 6.858  & 6.818  & \textbf{36.736} & 6.273  & 6.428  & \textbf{37.768} \\
\hline
\multirow{2}{*}{Office2} 
& mIoU $\uparrow$  & 6.810  & 6.912  & \textbf{40.952} & 8.450  & 8.836  & \textbf{46.701} \\
& mAcc $\uparrow$  & 16.760  & 18.337 & \textbf{88.098} & 29.269 & 29.729 & \textbf{92.155} \\
\hline
\multirow{2}{*}{Office3}   
& mIoU $\uparrow$  & 3.761  & 6.020  & \textbf{29.589} & 11.469 & 11.704 & \textbf{38.073} \\
& mAcc $\uparrow$  & 9.878  & 14.022 & \textbf{81.850} & 35.965 & 35.845 & \textbf{85.782} \\
\hline
\multirow{2}{*}{Office4} 
& mIoU $\uparrow$  & 10.124 & 11.310  & \textbf{32.496} & 13.609 & 14.179 & \textbf{44.225} \\
& mAcc $\uparrow$  & 21.978 & 19.523 & \textbf{56.657} & 28.891 & 29.890 & \textbf{73.817} \\
\hline
\multirow{2}{*}{Average} 
& mIoU $\uparrow$  & 6.025  & 6.838  & \textbf{27.146} & 9.160  & 9.358  & \textbf{32.679} \\
& mAcc $\uparrow$  & 17.237 & 18.864 & \textbf{66.994} & 27.106 & 27.454 & \textbf{69.277} \\
\hline
\Xhline{3\arrayrulewidth}
\end{tabular}
\end{adjustbox}
\end{table*}

\begin{table}[!t]
\renewcommand{\arraystretch}{1.1}
\caption{Performance comparison with other methods on segmentation results of novel views from the Replica dataset when initialized with fine-tuned CLIP models (refer to Section \ref{section:comparison_SOTA}). 
Best results are highlighted as \textbf{bold}. 
}
\label{table:SOTA_Replica_weakly}
\centering
\small
\begin{adjustbox} {width=.93\linewidth}
\begin{tabular}{ c c ||  p{0.9cm}<{\centering} p{0.9cm}<{\centering} | p{0.9cm}<{\centering} p{0.9cm}<{\centering}  }
\Xhline{3\arrayrulewidth}
\multirow{2}{*}{Dataset} & \multirow{2}{*}{Metric} &  {DFF} & {Ours} & {FCCLIP}  & {Ours} \\ 
~ & ~  & \cite{DFF} &  & \cite{FCCLIP}  &   \\
\hline\hline
\multirow{2}{*}{Room0}   
& mIoU $\uparrow$   & 25.289 & \textbf{30.233} & 36.322 & \textbf{42.875} \\
& mAcc $\uparrow$   & 68.400 & \textbf{71.830} & 75.648 & \textbf{81.491} \\
\hline
\multirow{2}{*}{Room1} 
& mIoU $\uparrow$   & 26.869 & \textbf{31.378} & 35.875 & \textbf{37.512} \\
& mAcc $\uparrow$   & 73.077 & \textbf{76.062} & 76.641 & \textbf{77.329} \\
\hline
\multirow{2}{*}{Room2}   
& mIoU $\uparrow$   & 27.892 & \textbf{28.489} & 32.782 & \textbf{35.276} \\
& mAcc $\uparrow$   & 80.314 & \textbf{80.973} & 83.794 & \textbf{84.670} \\
\hline
\multirow{2}{*}{Office0} 
& mIoU $\uparrow$   & 19.610 & \textbf{23.790} & 20.215 & \textbf{30.823} \\
& mAcc $\uparrow$   & 63.826 & \textbf{75.400} & 68.730 & \textbf{75.128} \\
\hline
\multirow{2}{*}{Office1}   
& mIoU $\uparrow$   & 14.250 & \textbf{25.329} & 19.252 & \textbf{25.449} \\
& mAcc $\uparrow$   & 51.297 & \textbf{60.107} & 50.129 & \textbf{50.592} \\
\hline
\multirow{2}{*}{Office2} 
& mIoU $\uparrow$   & 33.498 & \textbf{39.708} & 43.641 & \textbf{54.971} \\
& mAcc $\uparrow$   & 86.713 & \textbf{89.489} & 93.231 & \textbf{94.385} \\
\hline
\multirow{2}{*}{Office3}   
& mIoU $\uparrow$   & 27.380 & \textbf{36.473} & 33.679 & \textbf{40.545} \\
& mAcc $\uparrow$   & 83.404 & \textbf{86.209} & 87.732 & \textbf{88.814} \\
\hline
\multirow{2}{*}{Office4} 
& mIoU $\uparrow$   & 31.148 & \textbf{44.915} & 50.973 & \textbf{58.593} \\
& mAcc $\uparrow$   & 74.276 & \textbf{75.956} & 86.609 & \textbf{88.296} \\
\hline
\multirow{2}{*}{Average} 
& mIoU $\uparrow$   & 25.742 & \textbf{32.539} & 34.092 & \textbf{40.756} \\
& mAcc $\uparrow$   & 72.663 & \textbf{77.003} & 77.814 & \textbf{80.088} \\
\hline
\Xhline{3\arrayrulewidth}
\end{tabular}
\end{adjustbox}
\end{table}

\begin{table*}[t]
\renewcommand{\arraystretch}{1.1}
\caption{Performance comparison with other methods on segmentation results of novel views from the ScanNet dataset when initialized with vanilla CLIP models (refer to Section \ref{section:comparison_SOTA}). Best results are highlighted as \textbf{bold}.}
\label{table:SOTA_ScanNet}
\centering
\begin{adjustbox} {width=0.93\linewidth}
\begin{tabular}{ r | c | c c | c c | c c | c c | c c }
\Xhline{4\arrayrulewidth}
\multirow{2}{*}{Method} & \multirow{2}{*}{Setting}  
& \multicolumn{2}{c|}{Scene0004}  & \multicolumn{2}{c|}{Scene0389} & \multicolumn{2}{c|}{Scene0494} & \multicolumn{2}{c|}{Scene0693}  
& \multicolumn{2}{c}{Average}  \\
& & mIoU & mAcc & mIoU & mAcc & mIoU & mAcc & mIoU & mAcc & mIoU & mAcc  \\ 
\hline\hline
LERF   \cite{LERF}  & CLIP  & 7.137   & 12.139 & 20.102 & 62.025 & 17.624  & 35.079   & 16.531   & 51.933   & 15.349   & 40.294 \\ 
3DOVS  \cite{3DOVS} & CLIP  & 9.763   & 15.666 & 20.596 & 64.099 & 18.556  & 36.223   & 22.292   & 54.141   & 17.808   & 42.532 \\  
Ours    & CLIP
& \textbf{34.670} & \textbf{73.524} & \textbf{26.511} & \textbf{76.355}  
& \textbf{52.977} & \textbf{85.029} & \textbf{30.753} & \textbf{63.889} 
& \textbf{36.228} & \textbf{74.699}\\
\hline 
LERF   \cite{LERF}  & OpenCLIP    & 1.830   & 1.840  & 3.725  & 8.081  & 8.907  & 16.718 & 4.230  & 6.734  & 4.673  & 8.3433 \\
3DOVS  \cite{3DOVS} & OpenCLIP    & 2.724  & 2.521   & 5.749  & 21.497 & 13.235 & 24.164 & 6.294  & 10.001 & 7.001  & 14.546 \\
Ours   & OpenCLIP
& \textbf{42.059} & \textbf{85.217} & \textbf{25.035} & \textbf{78.948} 
& \textbf{56.833} & \textbf{88.210} & \textbf{25.129} & \textbf{50.616} 
& \textbf{37.264} & \textbf{75.748}\\ 
\Xhline{4\arrayrulewidth}
\end{tabular}
\end{adjustbox}
\end{table*}

\begin{table*}[t]
\renewcommand{\arraystretch}{1.1}
\caption{Performance comparison with other methods on segmentation results of novel views from the ScanNet dataset when initialized with fine-tuned CLIP models (refer to Section \ref{section:comparison_SOTA}). Best results are highlighted as \textbf{bold}. 
}
\label{table:SOTA_ScanNet_weakly}
\centering
\begin{adjustbox} {width=0.85\linewidth}
\begin{tabular}{ r | c c | c c | c c | c c | c c }
\Xhline{4\arrayrulewidth}
\multirow{2}{*}{Method} 
& \multicolumn{2}{c|}{Scene0004}  & \multicolumn{2}{c|}{Scene0389} & \multicolumn{2}{c|}{Scene0494} & \multicolumn{2}{c|}{Scene0693}  
& \multicolumn{2}{c}{Average}  \\
& mIoU & mAcc & mIoU & mAcc & mIoU & mAcc & mIoU & mAcc & mIoU & mAcc  \\ 
\hline\hline
DFF   \cite{DFF}    & 38.941   & 86.849   & 30.715   & 82.082   & 50.913   & 89.517   & 43.364   & 74.090   & 40.983   & 83.134  \\
Ours    
& \textbf{47.611} & \textbf{87.569} & \textbf{31.564} & \textbf{85.866} 
& \textbf{61.318} & \textbf{90.448} & \textbf{45.242} & \textbf{75.413} 
& \textbf{46.434} & \textbf{84.824} \\
\hline 
FC CLIP   \cite{FCCLIP}  & 53.093   & 88.258   & 33.000   & 85.696   & 56.149   & 85.393   & 46.645   & 82.424   & 47.222   & 85.443 \\
Ours   
& \textbf{55.329} & \textbf{89.817} & \textbf{41.290} & \textbf{87.490} 
& \textbf{60.904} & \textbf{89.672} & \textbf{51.465} & \textbf{83.144} 
& \textbf{52.247} & \textbf{87.531} \\ 
\Xhline{4\arrayrulewidth}
\end{tabular}
\end{adjustbox}
\end{table*}

\begin{table*}[t] 
\renewcommand{\arraystretch}{1.1}
\caption{
Performance comparison with other methods on segmentation results of novel views from the 3DOVS dataset when initialized with the same vanilla CLIP model. Best results are highlighted as \textbf{bold}. 
}
\label{table:SOTA_3DOVS}
\centering
\begin{adjustbox} {width=\linewidth}
\begin{tabular}{ r | c c | c c | c c | c c | c c | c c | c c }
\Xhline{4\arrayrulewidth}
\multirow{2}{*}{Method} 
& \multicolumn{2}{c|}{Bed}  & \multicolumn{2}{c|}{Sofa} & \multicolumn{2}{c|}{Lawn} & \multicolumn{2}{c|}{Room}  & \multicolumn{2}{c|}{Bench} & \multicolumn{2}{c|}{Table}   
& \multicolumn{2}{c}{Average}  \\
& mIoU & mAcc & mIoU & mAcc & mIoU & mAcc & mIoU & mAcc & mIoU & mAcc & mIoU & mAcc & mIoU & mAcc \\ 
\hline\hline
LERF   \cite{LERF}    & 73.5 & 86.9 & 27.0 & 43.8 & 73.7 & 93.5 & 46.6 & 79.8 & 53.2 & 79.7 & 33.4 & 41.0 & 51.2 & 70.8 \\ 
3DOVS   \cite{3DOVS}  & 89.5 & 96.7 & 74.0 & \textbf{91.6} & 88.2 & 97.3 & 92.8 & \textbf{98.9} & 89.3 & 96.3 & 88.8 & 96.5 & 87.1 & 96.2 \\ 
\hline 
Ours   
& \textbf{94.2}  & \textbf{98.9}  &  \textbf{85.7} & 91.5 & \textbf{96.9} & \textbf{99.4} & \textbf{93.3} & \textbf{98.9} & \textbf{95.8} & \textbf{98.9} & \textbf{89.8} & \textbf{97.2} & \textbf{92.6} & \textbf{97.5} \\ 
\Xhline{4\arrayrulewidth}
\end{tabular}
\end{adjustbox}
\end{table*}

\section{Experiments and Discussion}
In this section, we initially delineate the evaluation setup and implementation details. 
Subsequently, we thoroughly compare our approach with recent state-of-the-art (SOTA) models. 
Finally, we present ablation experiments and discussions to demonstrate the effectiveness of our proposed strategies.

\subsection{Experimental Setup}
\textbf{Evaluation Datasets.} 
To evaluate the effectiveness of our approach, we conduct experiments on two widely-used multi-view indoor scene datasets: Replica \cite{replica} and ScanNet \cite{scannet}. 
Replica contains diverse high-quality indoor scenes featuring rooms and offices, each characterized by photo-realistic textures, dense geometry, and per-primitive semantic classes. In our experiments, we utilize eight commonly used scenes from Replica, namely room0-2 and office0-4 \cite{replica}. 
ScanNet is a real-world indoor scene dataset that provides camera poses and semantic segmentation labels. The camera poses for each scene in ScanNet are obtained using BundleFusion \cite{Bundlefusion}. Our experiments on ScanNet focus on four representative scenes: Scene0004, Scene0389, Scene0494, and Scene0693. 

For each scene in the mentioned datasets, we employ uniform frame sampling from video sequences within the scene, generating a variable number of frames ranging from 200 to 300. 
Specifically, every scene in the Replica dataset consists of 300 images. Regarding ScanNet, Scene0004, Scene0389, Scene0494, and Scene0693 comprise 233, 236, 247, and 289 images, respectively. 
The image resolution for each scene in Replica is 640 $\times$ 480, while for ScanNet, it is 648 $\times$ 484. 
Following previous works LERF \cite{LERF} and 3DOVS \cite{3DOVS}, the images within each scene are partitioned into a training set and a testing set. 
In our experiment, we exclusively use every $10$-$th$ frame for testing, while the remaining frames are allocated to the training set. 
Consequently, RGB images and corresponding poses from the training set are employed for reconstruction and semantic field training. The ground-truth semantic labels from the testing set are utilized solely for evaluation purposes.

% To further assess the generalization of our approach, we evaluate our method on ten \textbf{in-facing} scenes from the 3DOVS \cite{3DOVS} and two outdoor scenes from the Mip-NeRF360 dataset \cite{mipnerf360}. 

In addition to conducting experiments in previous challenging ``inside-out" indoor scenarios, we further evaluate the generalization of our approach on six scenes utilized in 3DOVS \cite{3DOVS} and two commonly used multi-view outdoor scenes (i.e., Bicycle, Garden) from the Mip-NeRF360 dataset \cite{mipnerf360}. 
The scenes featured in 3DOVS consist of 28 to 37 images and are oriented face-forwarding, whose views are sampled in an ``outside-in" manner, indicating a certain amount of overlap across views. 
We follow the experimental protocol outlined in 3DOVS, which utilizes posed training images for semantic field optimization and semantic segmentation labels from the testing set for evaluation. 
Due to the absence of annotated segmentation maps for scenes in Mip-NeRF360, we randomly selected certain images for qualitative evaluations.

\textbf{Evaluation Metrics.} 
To assess the performance of our method and its counterparts, we adopt the mean Intersection-over-Union (mIoU) and mean pixel accuracy (mAcc).
% The definitions for these metrics are provided below. 

\subsection{Implementation Details} 
\textbf{Model Architecture. }
Following \cite{3DOVS}, we employ TensoRF \cite{tensorf} as the base NeRF architecture for efficiency. The plane size of TensoRF aligns with the original default setting. 
The RGB and CLIP feature branches within TensoRF share a common volume and utilize identical intermediate features, whereas the density volume remains independent.  Within TensoRF, the original MLP architecture in TensoRF is utilized to extract view-dependent RGB values. Additionally, another MLP is employed, excluding view direction input, to extract the rendered CLIP feature.
For the CLIP \cite{OpenAICLIP} and OpenCLIP \cite{OpenCLIP} models, we configure the image encoder and text encoder using the ViT-B/16 model settings to extract image and text features, respectively. 
For the DINO \cite{DINO} model, we apply version 1 dino$\_$vitb8 for DINO feature extraction. 
In the case of the SAM \cite{SAM} model, we employ the ViT-H model setting to generate region proposals. 
% Particularly, the CLIP model and SAM model are frozen during training. 

\textbf{Training. }
Our approach is implemented using PyTorch on an NVIDIA A100 GPU. 
Following the same training settings of \cite{3DOVS}, Adam optimizer with $betas = (0.9, 0.99)$ is employed to train our model for 15,000 iterations. 
The training batch contains 4096 rays. The learning rates for the volume and MLP branch are set to 2$e^{-2}$ and 1$e^{-4}$, respectively. A learning rate decay with a factor of 0.1 is applied for adjustment. 
Preceding the training process, the CLIP features and SAM's region proposals of training views are pre-computed offline. DINO features, on the other hand, are computed online during training. 
For a scene with a resolution of 640 $\times$ 480, our model is trained for approximately 1 hour.

\textbf{Rendering.} 
After training, OV-NeRF takes around 3.8 seconds to render an image at a resolution of 640 $\times$ 480.

\textbf{Hyperparameter.}
In our experiment, we set the update interval $\mathcal{N}$ = 1000 for pseudo label update. The $\mathcal{T}$ is set to 10,000 to start the cross-view self-enhancement strategy.

\subsection{Comparison with State-of-the-art Methods} \label{section:comparison_SOTA}
In this section, for a comprehensive evaluation, we compare our approach in two settings, which are categorized based on whether the CLIP model has been fine-tuned or not: \textit{I) vanilla CLIP} and\textit{ II) fine-tuned CLIP}. 
I) Vanilla CLIP involves using the original pre-trained CLIP models, such as CLIP \cite{OpenAICLIP} and OpenCLIP \cite{OpenCLIP}. 
II) Fine-tuned CLIP includes leveraging fine-tuned CLIP models, exemplified by LSeg \cite{LSeg}, which is fine-tuned on the 2D COCO \cite{coco} dataset. 
In the aforementioned categorization, LERF \cite{LERF} and 3DOVS \cite{3DOVS} distill knowledge from vanilla CLIP, placing them in category I. 
DFF utilizes features from the 2D open-vocabulary segmentation model LSeg \cite{LSeg}, categorizing it under category II. 
Furthermore, to facilitate a comprehensive comparison within category II, FC CLIP \cite{FCCLIP}, the state-of-the-art 2D open-vocabulary segmentation model, is adopted by distilling its segmentation results into our NeRF model for comparison. 
Note that all methods are trained on the same training set and validated on the same test set for fair comparisons.

\textbf{Quantitative Comparison.}
We present the quantitative results in Table \ref{table:SOTA_Replica}, \ref{table:SOTA_Replica_weakly}, \ref{table:SOTA_ScanNet}, \ref{table:SOTA_ScanNet_weakly}, and \ref{table:SOTA_3DOVS}.  
Our proposed approach consistently demonstrates superior performance across all scenes from various datasets and initialization settings. 
Specifically,  
1) in Table \ref{table:SOTA_Replica}, our approach exhibits a significant mIoU improvement of 20.308\% and 23.321\% mIoU for CLIP-based and OpenCLIP-based initializations over the second-best 3DOVS, respectively. 
These results affirm the robust and accurate performance of our method. 
2) From Tabel \ref{table:SOTA_Replica_weakly}, the performance gains with different fine-tuned CLIP model initializations showcase the effectiveness of our method. Notably, our method achieves better results with higher-quality initialization performance, indicating the efficacy and robustness of our strategy under various initializations. 
3) The similar and superior performance on ScanNet, as presented in Tables \ref{table:SOTA_ScanNet} and \ref{table:SOTA_ScanNet_weakly}, demonstrates the strong generalization ability of our method in real-world scenes.  
4) Furthermore, our approach showcases stable generalization performance and superior results when dealing with face-forwarding scenes, as shown in Table \ref{table:SOTA_3DOVS}. 
These quantitative results indicate a strong capability of our approach to take good advantage of single-view and cross-view information for accurate 3D open-vocabulary segmentation.

\begin{figure*}
\centering
\includegraphics[width=.95\linewidth]{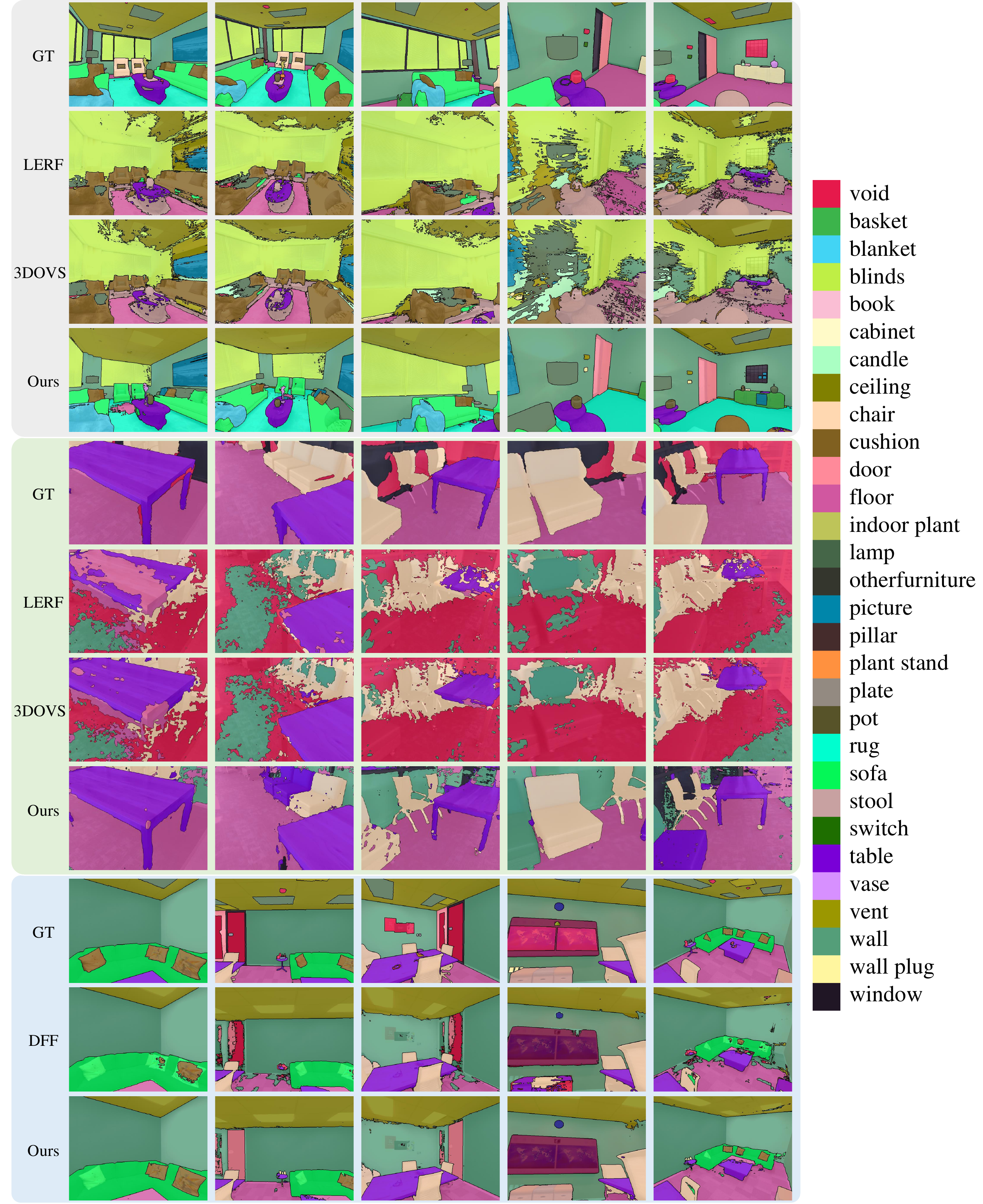}
\caption{
Qualitative results of various NeRF-based 3D open-vocabulary segmentation methods with different initialization, including CLIP model ($2^{nd} \sim 4^{th}$), OpenCLIP model ($6^{th} \sim 8^{th}$), and fine-tuned CLIP model ($10^{th} \sim 11^{th}$). Our method achieves more accurate and view-consistent results in various scenes. 
}
\label{fig:vis_comparisons}
\end{figure*}

\begin{figure}[!t]
\centering
\includegraphics[width=\linewidth]{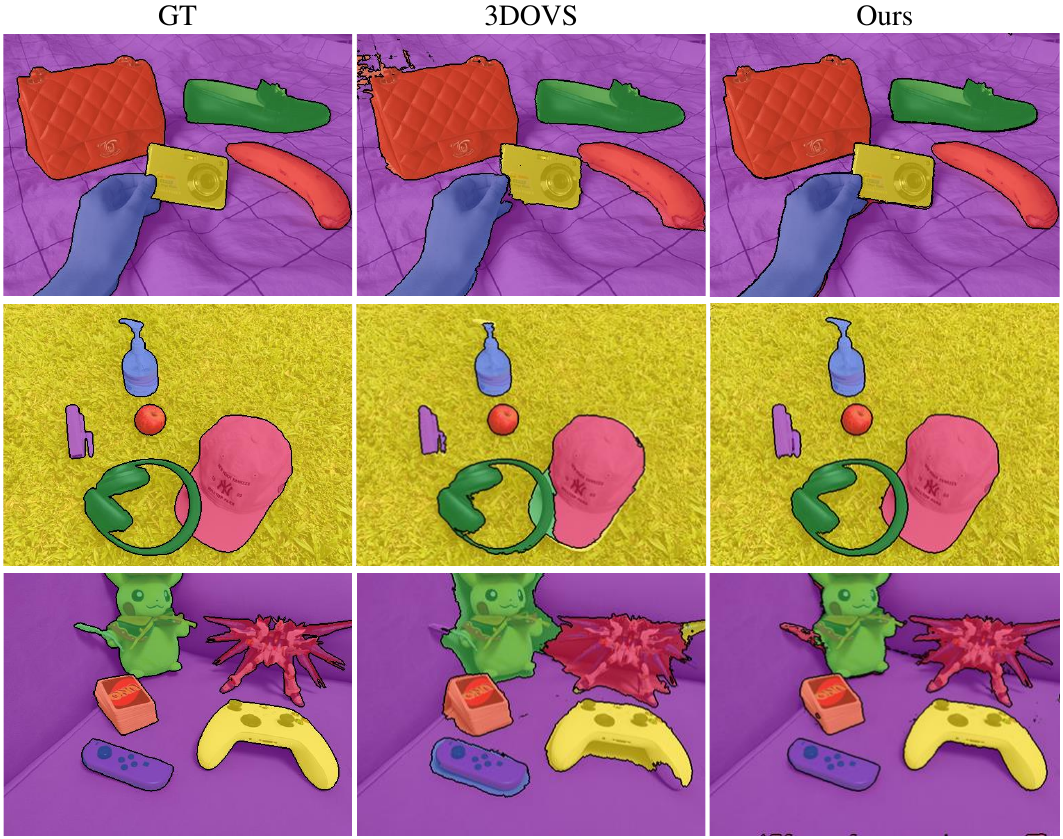}
\caption{
Visual open-vocabulary segmentation comparisons on face-forwarding scenes provided from 3DOVS \cite{3DOVS}. 
Our approach can produce more accurate results, especially with sharp object boundaries. 
}
\label{fig:vis_3dovs_scene}
\end{figure}

\begin{figure}[!t]
\centering
\includegraphics[width=.95\linewidth]{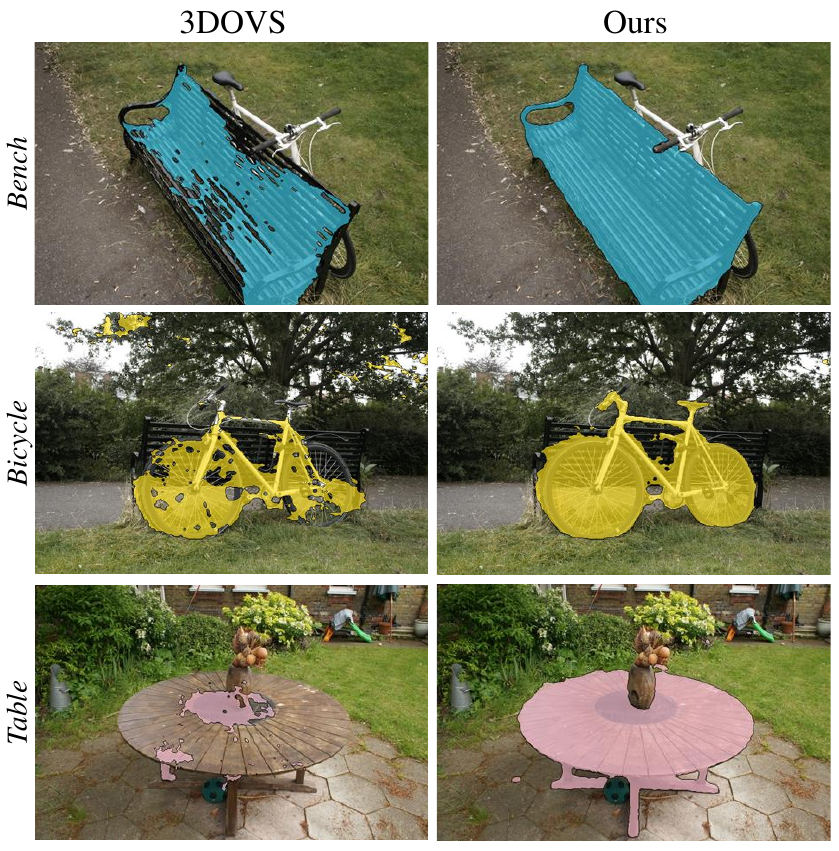}
\caption{
Visual open-vocabulary segmentation comparisons on outdoor scenes from the Mip-NeRF360 dataset \cite{mipnerf360}. 
Our method yields more precise results. 
}
\label{fig:vis_mipnerf}
\end{figure}

\textbf{Qualitative Comparison.}
We present the qualitative results produced by our method alongside comparisons with other approaches, including CLIP-based initialization, OpenCLIP-based initialization, and fine-tuned CLIP-based initialization, as illustrated in Fig. \ref{fig:vis_comparisons} and \ref{fig:vis_mipnerf}. 
We can observe that 
1) the existing SOTA methods struggle to accurately segment object boundaries, whereas our method consistently delivers precise segmentation results, such as the \textit{Blinds} in 2$^{nd}$ $\sim$ 4$^{th}$ rows. 
This precision is primarily attributed to our proposed Region Semantic Ranking Regularization, which refines object boundaries effectively by leveraging region masks derived from SAM. 
2) Our approach exhibits spatially continuous rendering results across different views, such as the \textit{Table} in 6$^{th}$ $\sim$ 8$^{th}$ rows. 
This achievement is credited to the Cross-view Self-enhancement strategy, allowing efficient utilization of 3D consistency constraints to enhance cross-view consistency.
3) As evident in the last two rows of Fig. \ref{fig:vis_comparisons}, we can see that our method can effectively reduce ambiguity, presenting more precise and view-consistent results. 
4) Additionally, as depicted in Fig. \ref{fig:vis_3dovs_scene} and \ref{fig:vis_mipnerf}, our method exhibits strong generalization ability by producing more accurate open-vocabulary segmentation results in face-forwarding and outdoor scenes, particularly showing sharp object boundaries. 
In summary, these qualitative results indicate that our proposed method produces accurate object boundaries, ensures superior 3D semantic consistency, and demonstrates robust performance.

\begin{table}[!t]
\renewcommand{\arraystretch}{1.2}
\caption{Ablation study for our approach with different strategies.}
\label{table_Ablation_all}
\centering
\small
\setlength{\tabcolsep}{0.65mm}{
\begin{tabular}{ c | c c | c | p{1.0cm}<{\centering} p{1.0cm}<{\centering} | p{1.0cm}<{\centering} p{1.0cm}<{\centering}}
\toprule[1pt] 
\multirow{2}{*}{Index} & \multicolumn{2}{c|}{Setting} & \multirow{2}{*}{Setting}  
& \multicolumn{2}{c|}{Room0} 
& \multicolumn{2}{c}{Scene0494} \\
& RSR & CSE  &  &  mIoU  & mAcc   & mIoU  & mAcc  \\
\midrule[0.8pt]
I        &            &               & CLIP  & 7.200  & 35.573 & 12.166 & 24.766 \\
II       & \checkmark &               & CLIP  & 19.098 & 60.907 & 49.161 & 82.051 \\
III      &            & \checkmark    & CLIP  & 26.780 & 70.467 & 50.762 & 82.486 \\
IV       & \checkmark & \checkmark    & CLIP  & \textbf{28.029} & \textbf{71.768} & \textbf{52.977} & \textbf{85.029} \\ 
\midrule[0.5pt]
V        &            &             & OpenCLIP  & 12.355 & 40.303 & 10.584 & 19.882 \\
VI       & \checkmark &             & OpenCLIP  & 26.577 & 75.563 & 50.268 & 82.820 \\
VII      &            & \checkmark  & OpenCLIP  & 35.109 & 81.213 & 53.494 & 85.618 \\
VIII     & \checkmark & \checkmark  & OpenCLIP  & \textbf{39.152} & \textbf{84.951} & \textbf{56.833} & \textbf{88.210} \\
\bottomrule[1pt]
\end{tabular}}
\end{table}

% \vspace*{-10pt}

\begin{table}[!t]
\renewcommand{\arraystretch}{1.2}
\caption{Ablation study for our proposed CSE with different settings. ``w/o CSE" denotes the model without using the CSE strategy during training. 
``w CSE$^*$" indicates that the outputs obtained from the learned semantic field are utilized without the RSR strategy. 
``$\mathcal{T}$" denotes the node that started using the CSE strategy during training. 
}
\label{table_Ablation_ST_start}
\centering
\small
\setlength{\tabcolsep}{0.65mm}
{\begin{tabular}{c | c | c | c c | c c }
\toprule[1pt]
\multirow{2}{*}{Index} & \multirow{2}{*}{Description} & \multirow{2}{*}{Setting}  & \multicolumn{2}{c|}{Room0} & \multicolumn{2}{c}{Scene0494} \\
&  &  &  mIoU  & mAcc   & mIoU  & mAcc  \\
\midrule[0.8pt]
1  & Ours w/o CSE     & CLIP  & 19.098 & 60.907 & 49.532 & 83.440 \\
2  & Ours w CSE$^*$     & CLIP  & 19.829 & 61.672 & 50.261 & 83.648 \\
3  & Ours ($\mathcal{T}$ = 8k)   & CLIP  & 27.763 & 71.311 & 51.275 & 83.838 \\
4  & Ours ($\mathcal{T}$ = 10k)  & CLIP  & \textbf{28.029} & \textbf{71.768} & \textbf{52.977} & \textbf{85.029} \\
5  & Ours ($\mathcal{T}$ = 12k)  & CLIP  & 26.634 & 69.788 & 51.180 & 83.899 \\
\midrule[0.5pt]
6  & Ours w/o CSE    & OpenCLIP  & 26.577 & 75.563 & 50.268 & 82.820 \\
7  & Ours w CSE$^*$    & OpenCLIP  & 27.999 & 75.851 & 54.279 & 86.219 \\
8  & Ours ($\mathcal{T}$ = 8k)  & OpenCLIP  & 36.942 & 82.604 & 54.295 & 86.827 \\
9 & Ours ($\mathcal{T}$ = 10k)  & OpenCLIP  & \textbf{39.152} & \textbf{84.951} & \textbf{56.833} & \textbf{88.210} \\ 
10  & Ours ($\mathcal{T}$ = 12k) & OpenCLIP  & 30.794 & 77.967 & 54.718 & 87.107 \\
\bottomrule[1pt]
\end{tabular}}
\end{table}

\begin{table}[!t]
\renewcommand{\arraystretch}{1.2}
\caption{Ablation study for ``$\mathcal{N}$", which denotes the update interval for pseudo label update in the proposed CSE strategy. 
}
\label{table_Ablation_ST_interval}
\centering
\small
\setlength{\tabcolsep}{0.75mm}
{\begin{tabular}{c | c | c | c c | c c }
\toprule[1pt]
\multirow{2}{*}{Index} & \multirow{2}{*}{Description} & \multirow{2}{*}{Setting}  & \multicolumn{2}{c|}{Room0} & \multicolumn{2}{c}{Scene0494} \\
&  &  &  mIoU  & mAcc   & mIoU  & mAcc  \\
\midrule[0.8pt]
a  & $\mathcal{N}$ = 5k     & CLIP  & 23.074 & 67.035 & 51.231 & 84.133 \\
b  & $\mathcal{N}$ = 2.5k   & CLIP  & 24.798 & 69.268 & 51.715 & 83.929 \\
c  & $\mathcal{N}$ = 1k     & CLIP  & \textbf{28.029} & \textbf{71.768} & \textbf{52.977} & \textbf{85.029} \\
d  & $\mathcal{N}$ = 0.5k   & CLIP  & 26.350 & 71.139 & 52.794 & 84.283 \\
\midrule[0.5pt]
e  & $\mathcal{N}$ = 5k     & OpenCLIP  & 33.929 & 80.244 & 50.528 & 84.187 \\
f  & $\mathcal{N}$ = 2.5k   & OpenCLIP  & 35.246 & 81.785 & 51.779 & 84.902 \\
g  & $\mathcal{N}$ = 1k     & OpenCLIP  & \textbf{39.152} & \textbf{84.951} & \textbf{56.833} & \textbf{88.210} \\ 
h  & $\mathcal{N}$ = 0.5k   & OpenCLIP  & 34.752 & 81.893 & 52.847 & 84.899 \\
\bottomrule[1pt]
\end{tabular}}
\end{table}

\subsection{Ablation Studies} \label{section:ablation_study}
In this section, we conduct ablation studies to validate the effectiveness of different components in our approach.

\begin{figure}[!t]
\centering
\includegraphics[width=\linewidth]{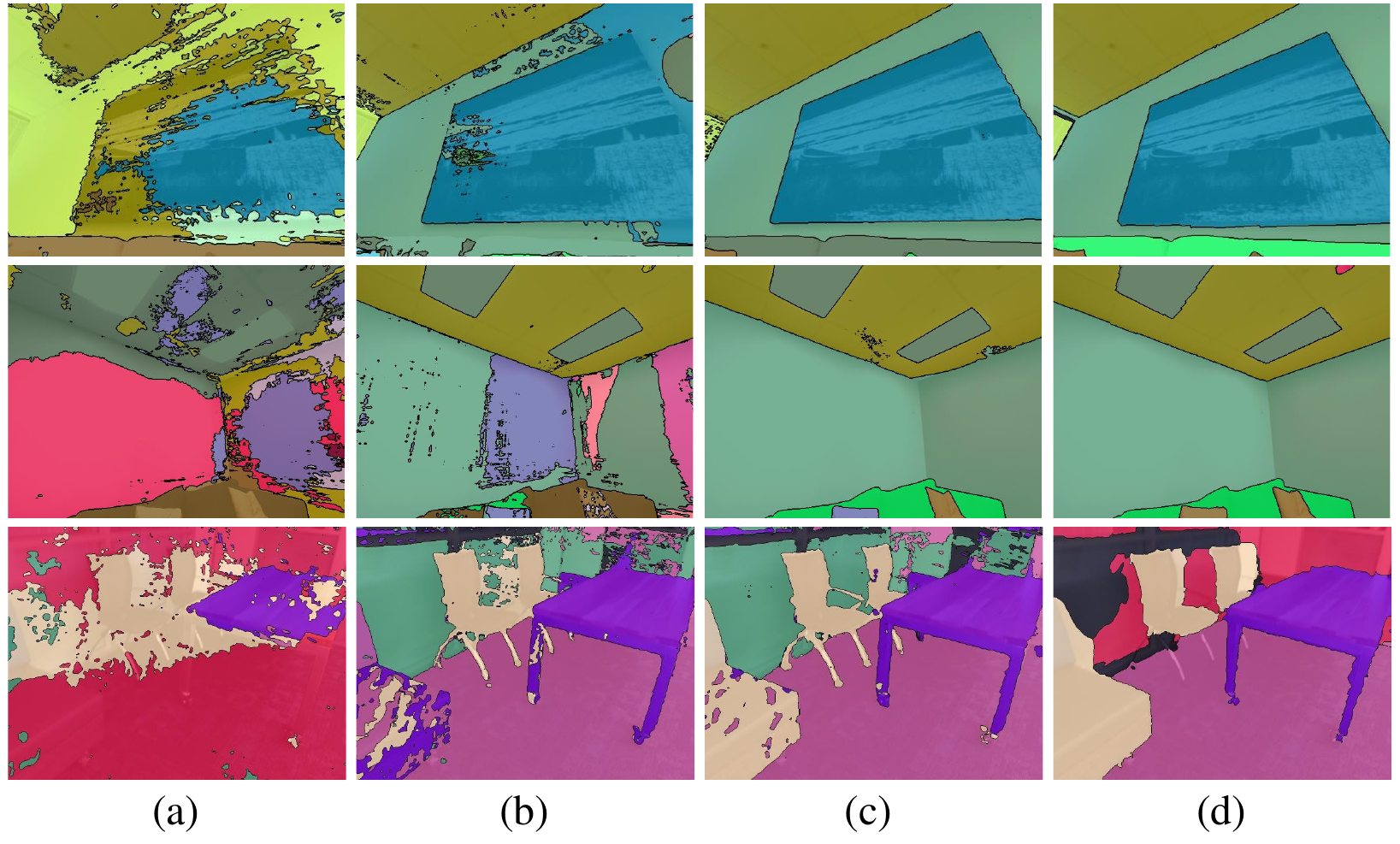}
\caption{
Visual results of ablation studies. (a) Baseline. (b) Baseline + RSR. (c) Baseline + RSR + CSE (OV-NeRF). (d) Ground Truth. 
}
\label{fig:vis_ablation}
\end{figure}

\textbf{Effectiveness of proposed components.}
We employ the 3DOVS \cite{3DOVS} and its semantic field optimization (equation (\ref{eq_3dovs_supervision})) as our baseline. 
As depicted in Table \ref{table_Ablation_all} and Fig. \ref{fig:vis_ablation}, we gradually apply our proposed strategies, including the Region Semantic Ranking (RSR) regularization and Cross-view Self-enhancement (CSE) strategy, to verify their effectiveness.

\begin{itemize}
    \item 
    Compared to the baseline, our RSR strategy, developed based on SAM, yields significant performance gains across various scenes and initialization settings. For instance, in the Room0 scene, RSR improves performance by 12.708\% and 14.222\% in terms of mIoU when using CLIP and OpenCLIP for initialization, respectively. 
    These results affirm that RSR rectifies coarse boundaries, promoting precise single-view relevancy map generation and accurate rendered results. 
    \item As evident in III and VII, the proposed CSE also effectively contributes to performance gains of 19.58\% and 22.754\% in terms of mIoU when using CLIP and OpenCLIP in the Room0 scene, respectively. 
    This phenomenon is attributed to Cross-view Self-enhancement (CSE) providing essential consistency constraints across multi-views, reducing ambiguity, and enhancing the quality of 3D semantic consistency. 
    \item Moreover, the cooperation of RSR and CSE, coupled with more precise single-view relevancy maps from RSR, consistently achieves better performance. This underscores the effectiveness of each component and highlights the availability of VLMs for NeRF-based 3D open-vocabulary semantic understanding. 
\end{itemize} 

\begin{figure}[!t]
\centering
\includegraphics[width=.95\linewidth]{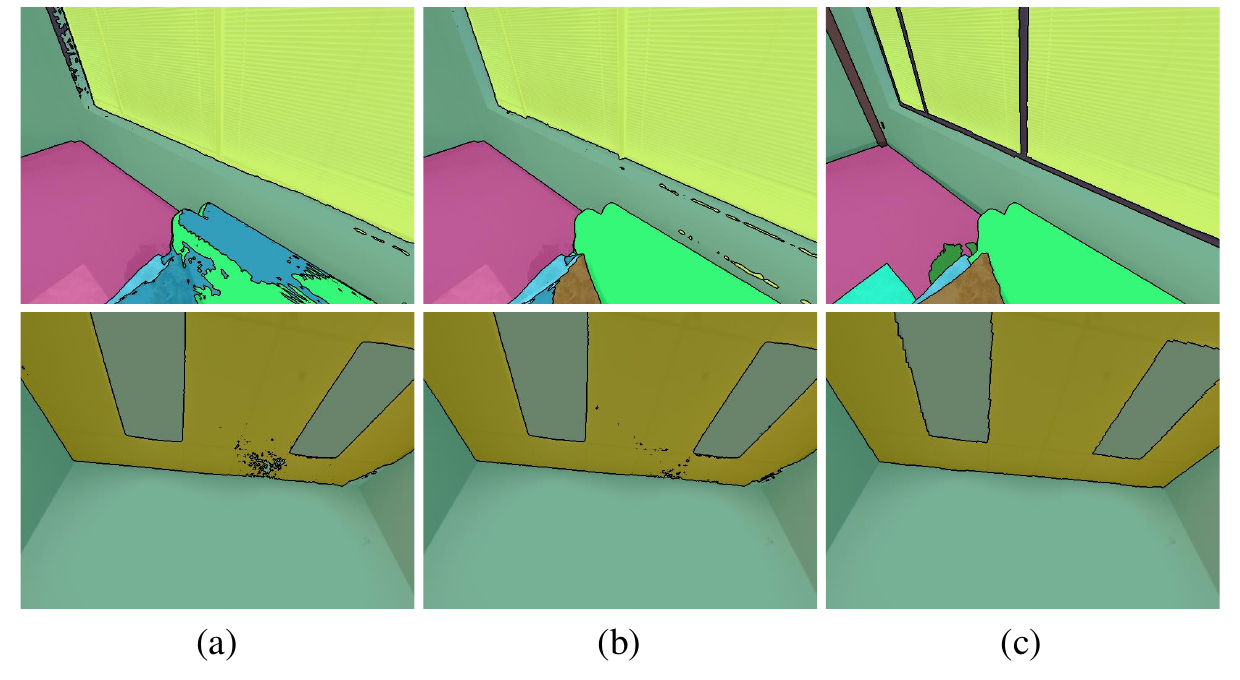}
\caption{
Visual ablation results for novel pseudo map synthesis (PMS). (a) OV-NeRF without PMS. (b) OV-NeRF. (c) Ground Truth. 
}
\label{fig:vis_ablation_nvs}
\end{figure}

Furthermore, we present visual results to show the effectiveness of each component in Fig. \ref{fig:vis_ablation}. 
Compared to the baseline, we can see that the RSR can enhance the integrity of objects as shown in (b), such as the picture in the first row and the table in the third row. 
Moreover, further applying our CSE strategy reduces the ambiguity of rendered results, as illustrated in (c). 
In addition, as illustrated in Fig. \ref{fig:vis_ablation_nvs}, leveraging the novel pseudo map synthesis process in CSE can provide valuable semantic information from extra unseen viewpoints to improve spatial continuity. 
In summary, these qualitative results demonstrate the effectiveness of our proposed strategies for accurate NeRF-based 3D open-vocabulary semantic understanding.

\textbf{Analysis of the $\mathcal{T}$ in CSE.}
To investigate the variant and the hyperparameter $\mathcal{T}$ in Cross-view Self-enhancement (CSE) strategy, we present comparison experiments in Table \ref{table_Ablation_ST_start}. 
Index 1 and Index 6, i.e., ``w/o CSE", denote the model without using the CSE strategy during training. 
Index 2 and Index 7, "w CSE$^*$," indicate that the outputs obtained from the learned semantic field are utilized without the Region Semantic Ranking (RSR) regularization. Particularly, in this configuration, the $\mathcal{T}$ is set to 10,000 to start using the outputs obtained from the learned semantic field. 
Index 3-5 and Index 8-10 represent the model leveraging the proposed CSE but utilizing different hyperparameters $\mathcal{T}$ to start CSE. 
Specifically, $\mathcal{T}=8k $ means that our model performs the cross-view self-enhancement loss $\mathcal{L}_{CSE}$ after 8,000 iterations of training. 

Compared to Indexes 1 and 6, the improvements observed in Indexes 2 and 7 demonstrate that after training a few iterations, the pseudo maps derived from the learned neural field surpass the original initial relevancy maps from CLIP models. 
These results validate the effectiveness of our CSE. 
Additionally, by applying RSR in CSE, the performances can be distinctly improved. 
This can be attributed to the fact that regional regularization further reduces the semantic ambiguity in the same region across different views. 
Meanwhile, we study the performance impact of when to implement CSE. 
We find that employing CSE either too early or too late, while both effective, represents suboptimal choices. When the $\mathcal{T}$ is set to 10k, the model consistently achieves superior results. 
Hence, the final configuration is set as $\mathcal{T}=10k$.

\textbf{Analysis of the update interval $\mathcal{N}$ in CSE.}
To investigate the impact of different update intervals in CSE, 
we conduct experiments with different settings in Table \ref{table_Ablation_ST_interval}.
A noticeable trend emerges as the interval decreases, signifying an increase in the frequency of updating and a corresponding steady improvement in performance. These experimental findings robustly validate the efficacy of cross-view self-enhancement training in consistently enhancing multi-view consistency at each stage. Consequently, it facilitates more precise view-consistent pseudo maps to enhance the network learning in subsequent stages. 
Nevertheless, excessively frequent updates might lead to performance degradation, potentially attributed to inadequate multi-view consistency training in each update. 
Therefore, considering these factors, we establish $\mathcal{N}=1k$ as the final setting in our experiments.

\subsection{Discussion on Real-world Applications and Impacts}
\textbf{Applications.} 
OV-NeRF enables precise open-vocabulary semantic understanding in real-world 3D scenes by leveraging multi-view images and language descriptions, eliminating the requirement for annotated semantic labels. 
This accurate and language-driven approach to 3D semantic understanding supports various applications, including robot navigation, robotic Grasping, and human-computer interaction. 
For instance, more accurate 3D scene understanding and object localization can improve the precision of spatial movement and object-grasping abilities in robots. Moreover, the elimination of labeling costs makes this process cost-effective. 
Additionally, modeling the 3D open-vocabulary neural radiance field empowers users to interact with and query 3D environments using natural language descriptions, offering a promising avenue for human-computer interaction. 
% 1) Accurate, 2) Cost-effective, 3) Interaction  

\textbf{Impacts.} 
OV-NeRF is a \textit{2D lifting} method that leverages language and visual knowledge from pre-trained 2D vision and language foundation models, such as CLIP and SAM, integrated with neural radiance fields to attain precise open-vocabulary 3D semantic understanding. 
This 2D lifting strategy eliminates the need for extensive annotated 3D datasets, offering a cost-efficient solution for 3D scene understanding tasks. 
In the future, we aspire for our work to stimulate more advancements in utilizing powerful 2D foundation models, e.g., Large Language Models (LLMs), for various 3D scene understanding tasks.

\section{Conclusion and Future Work}
In this paper, we present OV-NeRF to address the challenges of NeRF-based 3D semantic understanding, by leveraging the capacities from vision and language foundation models. In OV-NeRF, the proposed Region Semantic Ranking (RSR) regularization produces precise single-view relevancy maps to train the OV-NeRF and Cross-view Self-enhancement ensures view-consistent segmentation results. Experimental results demonstrate our method outperforms the SOTA approaches on synthetic and real-world benchmark datasets by a large margin, showing the superiority of our method. Additionally, our method consistently demonstrates superior performance across diverse CLIP configurations, affirming its generalizability.

\textbf{Limitations and Future works.} 
Although OV-NeRF capitalizes on the efficient TensoRF for constructing semantic fields, it encounters a common efficiency limitation shared by existing NeRF-based approaches, making real-time rendering challenging. This limitation stems from the inherent efficiency bottleneck in NeRF's ray-marching volume rendering technique. 
In the future, we will explore more efficient methods to accelerate rendering speed while preserving accuracy, especially tailored for processing large-scale scenes.

\ifCLASSOPTIONcaptionsoff
  \newpage
\fi

% \clearpage
\bibliographystyle{IEEEtran}
\bibliography{OV-NeRF}

% Generated by IEEEtran.bst, version: 1.14 (2015/08/26)
\begin{thebibliography}{10}
\providecommand{\url}[1]{#1}
\csname url@samestyle\endcsname
\providecommand{\newblock}{\relax}
\providecommand{\bibinfo}[2]{#2}
\providecommand{\BIBentrySTDinterwordspacing}{\spaceskip=0pt\relax}
\providecommand{\BIBentryALTinterwordstretchfactor}{4}
\providecommand{\BIBentryALTinterwordspacing}{\spaceskip=\fontdimen2\font plus
\BIBentryALTinterwordstretchfactor\fontdimen3\font minus \fontdimen4\font\relax}
\providecommand{\BIBforeignlanguage}[2]{{%
\expandafter\ifx\csname l@#1\endcsname\relax
\typeout{** WARNING: IEEEtran.bst: No hyphenation pattern has been}%
\typeout{** loaded for the language `#1'. Using the pattern for}%
\typeout{** the default language instead.}%
\else
\language=\csname l@#1\endcsname
\fi
#2}}
\providecommand{\BIBdecl}{\relax}
\BIBdecl

\bibitem{nerf}
B.~Mildenhall, P.~P. Srinivasan, M.~Tancik, J.~T. Barron, R.~Ramamoorthi, and R.~Ng, ``Nerf: Representing scenes as neural radiance fields for view synthesis,'' in \emph{Proceedings of the European Conference on Computer Vision}.\hskip 1em plus 0.5em minus 0.4em\relax Springer, 2020, pp. 405--421.

\bibitem{mipnerf}
J.~T. Barron, B.~Mildenhall, M.~Tancik, P.~Hedman, R.~Martin-Brualla, and P.~P. Srinivasan, ``Mip-nerf: A multiscale representation for anti-aliasing neural radiance fields,'' in \emph{Proceedings of the IEEE/CVF International Conference on Computer Vision}, 2021, pp. 5855--5864.

\bibitem{zipnerf}
J.~T. Barron, B.~Mildenhall, D.~Verbin, P.~P. Srinivasan, and P.~Hedman, ``Zip-nerf: Anti-aliased grid-based neural radiance fields,'' in \emph{Proceedings of the IEEE/CVF International Conference on Computer Vision}, 2023, pp. 19\,697--19\,705.

\bibitem{tensorf}
A.~Chen, Z.~Xu, A.~Geiger, J.~Yu, and H.~Su, ``Tensorf: Tensorial radiance fields,'' in \emph{Proceedings of the European Conference on Computer Vision}.\hskip 1em plus 0.5em minus 0.4em\relax Springer, 2022, pp. 333--350.

\bibitem{instantngp}
T.~M{\"u}ller, A.~Evans, C.~Schied, and A.~Keller, ``Instant neural graphics primitives with a multiresolution hash encoding,'' \emph{ACM Transactions on Graphics}, vol.~41, no.~4, pp. 1--15, 2022.

\bibitem{nerfreview}
K.~Gao, Y.~Gao, H.~He, D.~Lu, L.~Xu, and J.~Li, ``Nerf: Neural radiance field in 3d vision, a comprehensive review,'' \emph{arXiv preprint arXiv:2210.00379}, 2022.

\bibitem{zhou2023dynpoint}
K.~Zhou, J.-X. Zhong, S.~Shin, K.~Lu, Y.~Yang, A.~Markham, and N.~Trigoni, ``Dynpoint: Dynamic neural point for view synthesis,'' in \emph{Proceedings of the Advances in Neural Information Processing Systems}, 2023.

\bibitem{li2023representing}
S.~Li, Z.~Xia, and Q.~Zhao, ``Representing boundary-ambiguous scene online with scale-encoded cascaded grids and radiance field deblurring,'' \emph{IEEE Transactions on Circuits and Systems for Video Technology}, 2023.

\bibitem{guo2024depth}
S.~Guo, Q.~Wang, Y.~Gao, R.~Xie, L.~Li, F.~Zhu, and L.~Song, ``Depth-guided robust point cloud fusion nerf for sparse input views,'' \emph{IEEE Transactions on Circuits and Systems for Video Technology}, 2024.

\bibitem{robotnavigation}
C.~Huang, O.~Mees, A.~Zeng, and W.~Burgard, ``Visual language maps for robot navigation,'' in \emph{Proceedings of the IEEE International Conference on Robotics and Automation}, 2023, pp. 10\,608--10\,615.

\bibitem{jaritz2019multi}
M.~Jaritz, J.~Gu, and H.~Su, ``Multi-view pointnet for 3d scene understanding,'' in \emph{Proceedings of the IEEE/CVF International Conference on Computer Vision Workshop}, 2019, pp. 3995--4003.

\bibitem{feng2020deep}
D.~Feng, C.~Haase-Sch{\"u}tz, L.~Rosenbaum, H.~Hertlein, C.~Glaeser, F.~Timm, W.~Wiesbeck, and K.~Dietmayer, ``Deep multi-modal object detection and semantic segmentation for autonomous driving: Datasets, methods, and challenges,'' \emph{IEEE Transactions on Intelligent Transportation Systems}, vol.~22, no.~3, pp. 1341--1360, 2020.

\bibitem{conceptfusion}
K.~M. Jatavallabhula, A.~Kuwajerwala, Q.~Gu, M.~Omama, G.~Iyer, S.~Saryazdi, T.~Chen, A.~Maalouf, S.~Li, N.~V. Keetha, A.~Tewari, J.~Tenenbaum, C.~de~Melo, M.~Krishna, L.~Paull, F.~Shkurti, and A.~Torralba, ``Conceptfusion: Open-set multimodal 3d mapping,'' in \emph{Proceedings of Robotics: Science and Systems}, 2023.

\bibitem{rong2021active}
M.~Rong, H.~Cui, Z.~Hu, H.~Jiang, H.~Liu, and S.~Shen, ``Active learning based 3d semantic labeling from images and videos,'' \emph{IEEE Transactions on Circuits and Systems for Video Technology}, vol.~32, no.~12, pp. 8101--8115, 2021.

\bibitem{yin2023dcnet}
F.~Yin, Z.~Huang, T.~Chen, G.~Luo, G.~Yu, and B.~Fu, ``Dcnet: Large-scale point cloud semantic segmentation with discriminative and efficient feature aggregation,'' \emph{IEEE Transactions on Circuits and Systems for Video Technology}, 2023.

\bibitem{shi2023temporal}
H.~Shi, R.~Li, F.~Liu, and G.~Lin, ``Temporal feature matching and propagation for semantic segmentation on 3d point cloud sequences,'' \emph{IEEE Transactions on Circuits and Systems for Video Technology}, 2023.

\bibitem{semanticNeRF}
S.~Zhi, T.~Laidlow, S.~Leutenegger, and A.~J. Davison, ``In-place scene labelling and understanding with implicit scene representation,'' in \emph{Proceedings of the IEEE/CVF International Conference on Computer Vision}, 2021, pp. 15\,838--15\,847.

\bibitem{OpenAICLIP}
A.~Radford, J.~W. Kim, C.~Hallacy, A.~Ramesh, G.~Goh, S.~Agarwal, G.~Sastry, A.~Askell, P.~Mishkin, J.~Clark \emph{et~al.}, ``Learning transferable visual models from natural language supervision,'' in \emph{Proceedings of the International conference on machine learning}.\hskip 1em plus 0.5em minus 0.4em\relax PMLR, 2021, pp. 8748--8763.

\bibitem{OpenCLIP}
M.~Cherti, R.~Beaumont, R.~Wightman, M.~Wortsman, G.~Ilharco, C.~Gordon, C.~Schuhmann, L.~Schmidt, and J.~Jitsev, ``Reproducible scaling laws for contrastive language-image learning,'' in \emph{Proceedings of the IEEE/CVF Conference on Computer Vision and Pattern Recognition}, 2023, pp. 2818--2829.

\bibitem{LSeg}
B.~Li, K.~Q. Weinberger, S.~Belongie, V.~Koltun, and R.~Ranftl, ``Language-driven semantic segmentation,'' in \emph{Proceedings of the International Conference on Learning Representations}, 2022.

\bibitem{FCCLIP}
Q.~Yu, J.~He, X.~Deng, X.~Shen, and L.-C. Chen, ``Convolutions die hard: Open-vocabulary segmentation with single frozen convolutional clip,'' in \emph{Proceedings of the Advances in Neural Information Processing Systems}, 2023.

\bibitem{regionclip}
Y.~Zhong, J.~Yang, P.~Zhang, C.~Li, N.~Codella, L.~H. Li, L.~Zhou, X.~Dai, L.~Yuan, Y.~Li \emph{et~al.}, ``Regionclip: Region-based language-image pretraining,'' in \emph{Proceedings of the IEEE/CVF Conference on Computer Vision and Pattern Recognition}, 2022, pp. 16\,793--16\,803.

\bibitem{zhou2022extract}
C.~Zhou, C.~C. Loy, and B.~Dai, ``Extract free dense labels from clip,'' in \emph{Proceedings of the European Conference on Computer Vision}.\hskip 1em plus 0.5em minus 0.4em\relax Springer, 2022, pp. 696--712.

\bibitem{CLIPSelf}
S.~Wu, W.~Zhang, L.~Xu, S.~Jin, X.~Li, W.~Liu, and C.~C. Loy, ``Clipself: Vision transformer distills itself for open-vocabulary dense prediction,'' in \emph{Proceedings of the International Conference on Learning Representations}, 2024.

\bibitem{Segclip}
H.~Luo, J.~Bao, Y.~Wu, X.~He, and T.~Li, ``Segclip: Patch aggregation with learnable centers for open-vocabulary semantic segmentation,'' in \emph{International Conference on Machine Learning}.\hskip 1em plus 0.5em minus 0.4em\relax PMLR, 2023, pp. 23\,033--23\,044.

\bibitem{xu2023side}
M.~Xu, Z.~Zhang, F.~Wei, H.~Hu, and X.~Bai, ``Side adapter network for open-vocabulary semantic segmentation,'' in \emph{Proceedings of the IEEE/CVF Conference on Computer Vision and Pattern Recognition}, 2023, pp. 2945--2954.

\bibitem{liang2023open}
F.~Liang, B.~Wu, X.~Dai, K.~Li, Y.~Zhao, H.~Zhang, P.~Zhang, P.~Vajda, and D.~Marculescu, ``Open-vocabulary semantic segmentation with mask-adapted clip,'' in \emph{Proceedings of the IEEE/CVF Conference on Computer Vision and Pattern Recognition}, 2023, pp. 7061--7070.

\bibitem{zhang2023simple}
H.~Zhang, F.~Li, X.~Zou, S.~Liu, C.~Li, J.~Yang, and L.~Zhang, ``A simple framework for open-vocabulary segmentation and detection,'' in \emph{Proceedings of the IEEE/CVF International Conference on Computer Vision}, 2023, pp. 1020--1031.

\bibitem{LERF}
J.~Kerr, C.~M. Kim, K.~Goldberg, A.~Kanazawa, and M.~Tancik, ``Lerf: Language embedded radiance fields,'' in \emph{Proceedings of the IEEE/CVF International Conference on Computer Vision}, 2023, pp. 19\,729--19\,739.

\bibitem{3DOVS}
K.~Liu, F.~Zhan, J.~Zhang, M.~Xu, Y.~Yu, A.~El~Saddik, C.~Theobalt, E.~Xing, and S.~Lu, ``Weakly supervised 3d open-vocabulary segmentation,'' in \emph{Proceedings of the Advances in Neural Information Processing Systems}, 2023, pp. 53\,433--53\,456.

\bibitem{SAM}
A.~Kirillov, E.~Mintun, N.~Ravi, H.~Mao, C.~Rolland, L.~Gustafson, T.~Xiao, S.~Whitehead, A.~C. Berg, W.-Y. Lo, P.~Dollar, and R.~Girshick, ``Segment anything,'' in \emph{Proceedings of the IEEE/CVF International Conference on Computer Vision}, 2023, pp. 4015--4026.

\bibitem{replica}
J.~Straub, T.~Whelan, L.~Ma, Y.~Chen, E.~Wijmans, S.~Green, J.~J. Engel, R.~Mur-Artal, C.~Ren, S.~Verma \emph{et~al.}, ``The replica dataset: A digital replica of indoor spaces,'' \emph{arXiv preprint arXiv:1906.05797}, 2019.

\bibitem{scannet}
A.~Dai, A.~X. Chang, M.~Savva, M.~Halber, T.~Funkhouser, and M.~Nie{\ss}ner, ``Scannet: Richly-annotated 3d reconstructions of indoor scenes,'' in \emph{Proceedings of the IEEE conference on computer vision and pattern recognition}, 2017, pp. 5828--5839.

\bibitem{Nerfies}
K.~Park, U.~Sinha, J.~T. Barron, S.~Bouaziz, D.~B. Goldman, S.~M. Seitz, and R.~Martin-Brualla, ``Nerfies: Deformable neural radiance fields,'' in \emph{Proceedings of the IEEE/CVF International Conference on Computer Vision}, 2021, pp. 5865--5874.

\bibitem{pixelnerf}
A.~Yu, V.~Ye, M.~Tancik, and A.~Kanazawa, ``pixelnerf: Neural radiance fields from one or few images,'' in \emph{Proceedings of the IEEE/CVF Conference on Computer Vision and Pattern Recognition}, 2021, pp. 4578--4587.

\bibitem{mipnerf360}
J.~T. Barron, B.~Mildenhall, D.~Verbin, P.~P. Srinivasan, and P.~Hedman, ``Mip-nerf 360: Unbounded anti-aliased neural radiance fields,'' in \emph{Proceedings of the IEEE/CVF Conference on Computer Vision and Pattern Recognition}, 2022, pp. 5470--5479.

\bibitem{Nerfren}
Y.-C. Guo, D.~Kang, L.~Bao, Y.~He, and S.-H. Zhang, ``Nerfren: Neural radiance fields with reflections,'' in \emph{Proceedings of the IEEE/CVF Conference on Computer Vision and Pattern Recognition}, 2022, pp. 18\,409--18\,418.

\bibitem{wang2022nerfcap}
K.~Wang, S.~Peng, X.~Zhou, J.~Yang, and G.~Zhang, ``Nerfcap: Human performance capture with dynamic neural radiance fields,'' \emph{IEEE Transactions on Visualization and Computer Graphics}, 2022.

\bibitem{Nerfplayer}
L.~Song, A.~Chen, Z.~Li, Z.~Chen, L.~Chen, J.~Yuan, Y.~Xu, and A.~Geiger, ``Nerfplayer: A streamable dynamic scene representation with decomposed neural radiance fields,'' \emph{IEEE Transactions on Visualization and Computer Graphics}, vol.~29, no.~5, pp. 2732--2742, 2023.

\bibitem{wei2023depth}
Y.~Wei, S.~Liu, J.~Zhou, and J.~Lu, ``Depth-guided optimization of neural radiance fields for indoor multi-view stereo,'' \emph{IEEE Transactions on Pattern Analysis and Machine Intelligence}, 2023.

\bibitem{Structnerf}
Z.~Chen, C.~Wang, Y.-C. Guo, and S.-H. Zhang, ``Structnerf: Neural radiance fields for indoor scenes with structural hints,'' \emph{IEEE Transactions on Pattern Analysis and Machine Intelligence}, 2023.

\bibitem{cp_decomposition}
J.~D. Carroll and J.-J. Chang, ``Analysis of individual differences in multidimensional scaling via an n-way generalization of “eckart-young” decomposition,'' \emph{Psychometrika}, vol.~35, no.~3, pp. 283--319, 1970.

\bibitem{n3f}
V.~Tschernezki, I.~Laina, D.~Larlus, and A.~Vedaldi, ``Neural feature fusion fields: 3d distillation of self-supervised 2d image representations,'' in \emph{Proceedings of the International Conference on 3D Vision}, 2022, pp. 443--453.

\bibitem{DINO}
M.~Caron, H.~Touvron, I.~Misra, H.~J{\'e}gou, J.~Mairal, P.~Bojanowski, and A.~Joulin, ``Emerging properties in self-supervised vision transformers,'' in \emph{Proceedings of the IEEE/CVF international conference on computer vision}, 2021, pp. 9650--9660.

\bibitem{nerfsos}
Z.~Fan, P.~Wang, Y.~Jiang, X.~Gong, D.~Xu, and Z.~Wang, ``Nerf-sos: Any-view self-supervised object segmentation on complex scenes,'' in \emph{Proceedings of the International Conference on Learning Representations}, 2022.

\bibitem{DFF}
S.~Kobayashi, E.~Matsumoto, and V.~Sitzmann, ``Decomposing nerf for editing via feature field distillation,'' in \emph{Proceedings of the Advances in Neural Information Processing Systems}, vol.~35, 2022, pp. 23\,311--23\,330.

\bibitem{flamingo}
J.-B. Alayrac, J.~Donahue, P.~Luc, A.~Miech, I.~Barr, Y.~Hasson, K.~Lenc, A.~Mensch, K.~Millican, M.~Reynolds \emph{et~al.}, ``Flamingo: a visual language model for few-shot learning,'' in \emph{Proceedings of the Advances in Neural Information Processing Systems}, vol.~35, 2022, pp. 23\,716--23\,736.

\bibitem{samtrack}
Y.~Cheng, L.~Li, Y.~Xu, X.~Li, Z.~Yang, W.~Wang, and Y.~Yang, ``Segment and track anything,'' \emph{arXiv preprint arXiv:2305.06558}, 2023.

\bibitem{sammedical}
J.~Ma, Y.~He, F.~Li, L.~Han, C.~You, and B.~Wang, ``Segment anything in medical images,'' \emph{Nature Communications}, vol.~15, no.~1, p. 654, 2024.

\bibitem{samhq}
L.~Ke, M.~Ye, M.~Danelljan, Y.~Liu, Y.-W. Tai, C.-K. Tang, and F.~Yu, ``Segment anything in high quality,'' in \emph{Proceedings of the Advances in Neural Information Processing Systems}, 2023.

\bibitem{Bundlefusion}
A.~Dai, M.~Nie{\ss}ner, M.~Zollh{\"o}fer, S.~Izadi, and C.~Theobalt, ``Bundlefusion: Real-time globally consistent 3d reconstruction using on-the-fly surface reintegration,'' \emph{ACM Transactions on Graphics}, vol.~36, no.~4, p.~1, 2017.

\bibitem{coco}
T.-Y. Lin, M.~Maire, S.~Belongie, J.~Hays, P.~Perona, D.~Ramanan, P.~Doll{\'a}r, and C.~L. Zitnick, ``Microsoft coco: Common objects in context,'' in \emph{Proceedings of the European Conference on Computer Vision}.\hskip 1em plus 0.5em minus 0.4em\relax Springer, 2014, pp. 740--755.

\end{thebibliography}

\begin{IEEEbiography}[{\includegraphics[width=1in,height=1.25in,clip,keepaspectratio]{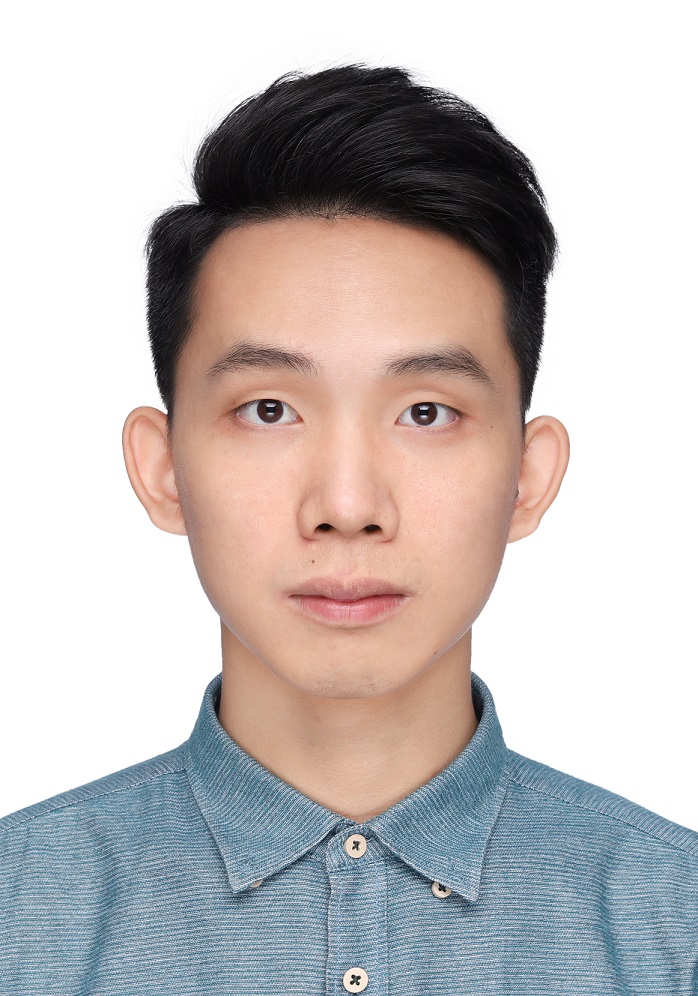}}]{Guibiao Liao}
received the B.E. degree from Tianjin University, China, in 2020. He is currently pursuing the Ph.D. degree at the School of Electronic and Computer Engineering, Peking University, China. 
His research interests include 3D computer vision and deep learning. 
\end{IEEEbiography}

\begin{IEEEbiography}[{\includegraphics[width=1in,height=1.25in,clip,keepaspectratio]{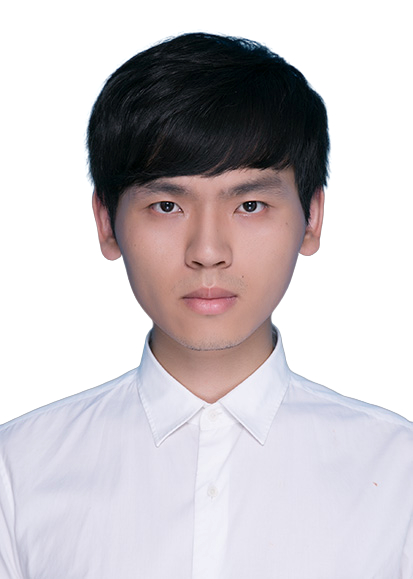}}]{Kaichen Zhou}
is presently working towards his Ph.D. in Computer Science at the University of Oxford’s Department of Computer Science. Before this, he earned his degree from the Department of Computing at Imperial College London. He has a keen interest in computer vision and robotics.
\end{IEEEbiography}
% \vspace{-15 mm}

\begin{IEEEbiography}[{\includegraphics[width=1in,height=1.25in,clip,keepaspectratio]{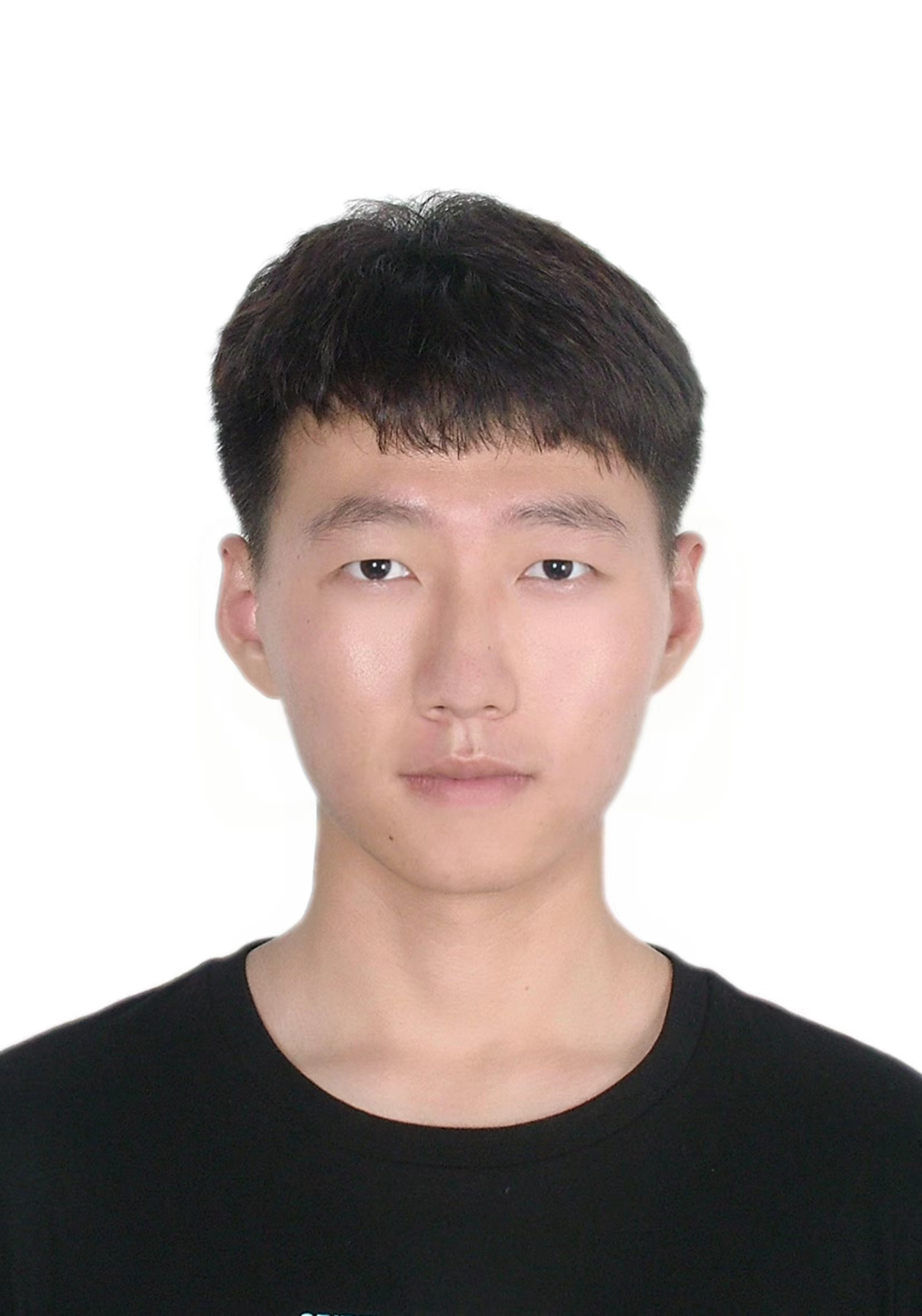}}]{Zhenyu Bao}
received the B.E. degree from the Department of Microelectronic Science and Engineering, Jilin University, in 2019, and the M.S. degree from the School of Electronic and Computer Engineering, Peking University, in 2022. He is currently pursuing the Ph.D. degree with the School of Computer Science from Peking University. His research interests lie in the domain of Novel View Synthesis and its applications in AR/XR. 
\end{IEEEbiography}
% \vspace{-70 mm}

\begin{IEEEbiography}[{\includegraphics[width=1in,height=1.25in,clip,keepaspectratio]{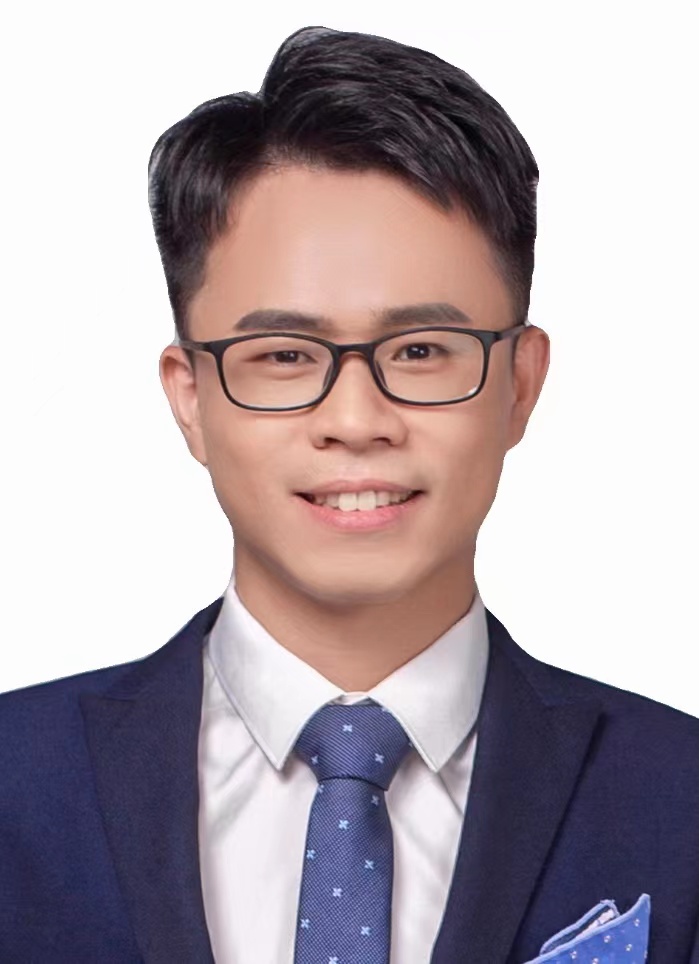}}]{Kanglin Liu} received the B.Eng. and Ph.D. degrees from Chongqing University, China, in 2013 and 2018, respectively. From 2016 to 2017, he was a Visiting Scholar at The University of Tennessee and the Oak Ridge National Laboratory. From 2018 to 2020, he was a Postdoctoral Fellow at Shenzhen University and the University of Nottingham. He is currently an Assistant Research Fellow with the Pengcheng Laboratory, in China.
His current interests include 3D reconstruction, neural rendering, and head avatar creation.
\end{IEEEbiography}
% \vspace{-70 mm}

\begin{IEEEbiography}[{\includegraphics[width=1in,height=1.25in,clip,keepaspectratio]{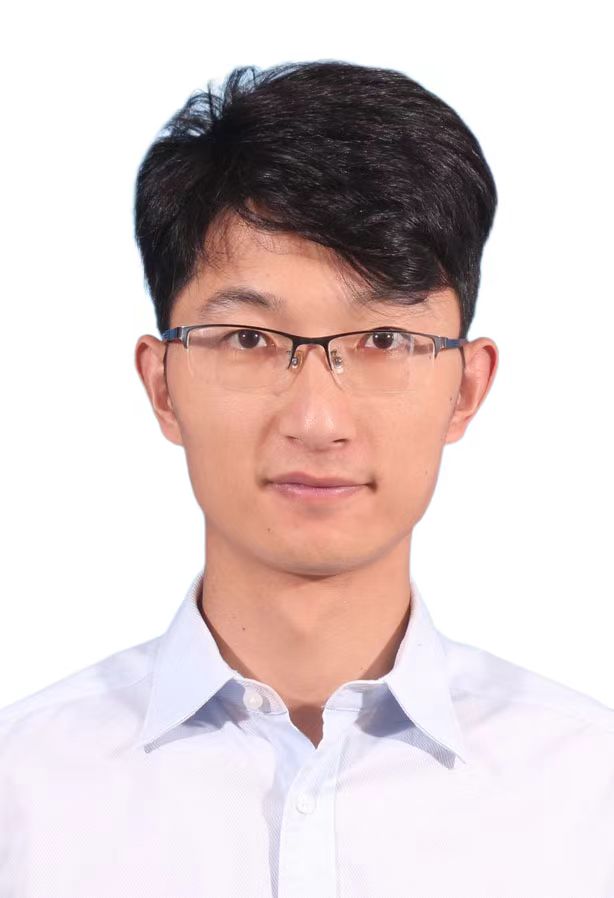}}]{Qing Li}
received the Ph.D. degree in computer vision from the University of Nottingham UK, in 2020. He is currently an assistant researcher in Pengcheng Laboratory China. His research interests include image-based indoor localization, visual SLAM, image processing, 3D reconstruction, and remote image interpretation. 
\end{IEEEbiography}

\end{document}